\pdfoutput=1
\def\ArxivMode{1}

\documentclass[10pt,twocolumn,letterpaper]{article}

\usepackage[pagenumbers]{cvpr}

\usepackage{microtype}

\usepackage{tabularx,array,multirow}
\usepackage{tikz}
\usetikzlibrary{arrows.meta}
\providecommand{\supplref}[2]{\ifdefined\ArxivMode #2\else #1\fi}
\providecommand{\mainref}[2]{\ifdefined\ArxivMode #2\else #1\fi}

\usepackage{colortbl}
\usepackage{algorithm}
\usepackage{algpseudocode}

\definecolor{maskYes}{HTML}{D8EFE0}
\definecolor{maskNo}{HTML}{F5F7FA}
\definecolor{maskSelf}{HTML}{DBEAF8}
\definecolor{maskCond}{HTML}{F7E5C6}
\definecolor{boxBg}{HTML}{F8FAFC}
\definecolor{boxBorder}{HTML}{94A8BC}
\definecolor{accentBlue}{HTML}{1D4770}
\definecolor{accentGreen}{HTML}{1B5434}
\definecolor{accentAmber}{HTML}{7A4E10}
\definecolor{accentPurple}{HTML}{4A2F70}

\algrenewcommand\algorithmicrequire{\textbf{Input:}}
\algrenewcommand\algorithmicensure{\textbf{Output:}}

\definecolor{cvprblue}{rgb}{0.21,0.49,0.74}
\usepackage[pagebackref,breaklinks,colorlinks,allcolors=cvprblue]{hyperref}

\def\confName{CVPR}
\def\confYear{2027}

\title{ContextRenderer: Dynamic-Context Video Generation\\for Interactive Game Worlds}

\author{%
Runze Wang$^{1*}$\quad
Fei Zha$^{1*}$\quad
Bing He$^{2*}$\quad
Ke Xu$^{1}$\quad
Zhongyuan Wu$^{3}$\\[0.12em]
Yuxuan Xie$^{4}$\quad
Wenzheng Wu$^{1}$\quad
Zhongyuan Liu$^{1\dagger}$\quad
Youcheng Cai$^{1\dagger}$\quad
Ligang Liu$^{1}$\\[0.3em]
{\normalsize $^{1}$University of Science and Technology of China\qquad $^{2}$Shanghai Jiao Tong University}\\
{\normalsize $^{3}$University of British Columbia\qquad $^{4}$Harbin Institute of Technology}\\[0.2em]
{\small $^*$Equal contribution\quad $^\dagger$Corresponding authors}
}

\makeatletter
\patchcmd{\@maketitle}{\vskip .375in}{\vskip .13in}{}{}
\patchcmd{\@maketitle}{\vspace*{24pt}}{\vspace*{8pt}}{}{}
\patchcmd{\@maketitle}{\vspace*{12pt}}{\vspace*{2pt}}{}{}
\apptocmd{\@maketitle}{\par\noindent
\begin{minipage}{\textwidth}
\centering
\includegraphics[width=\textwidth]{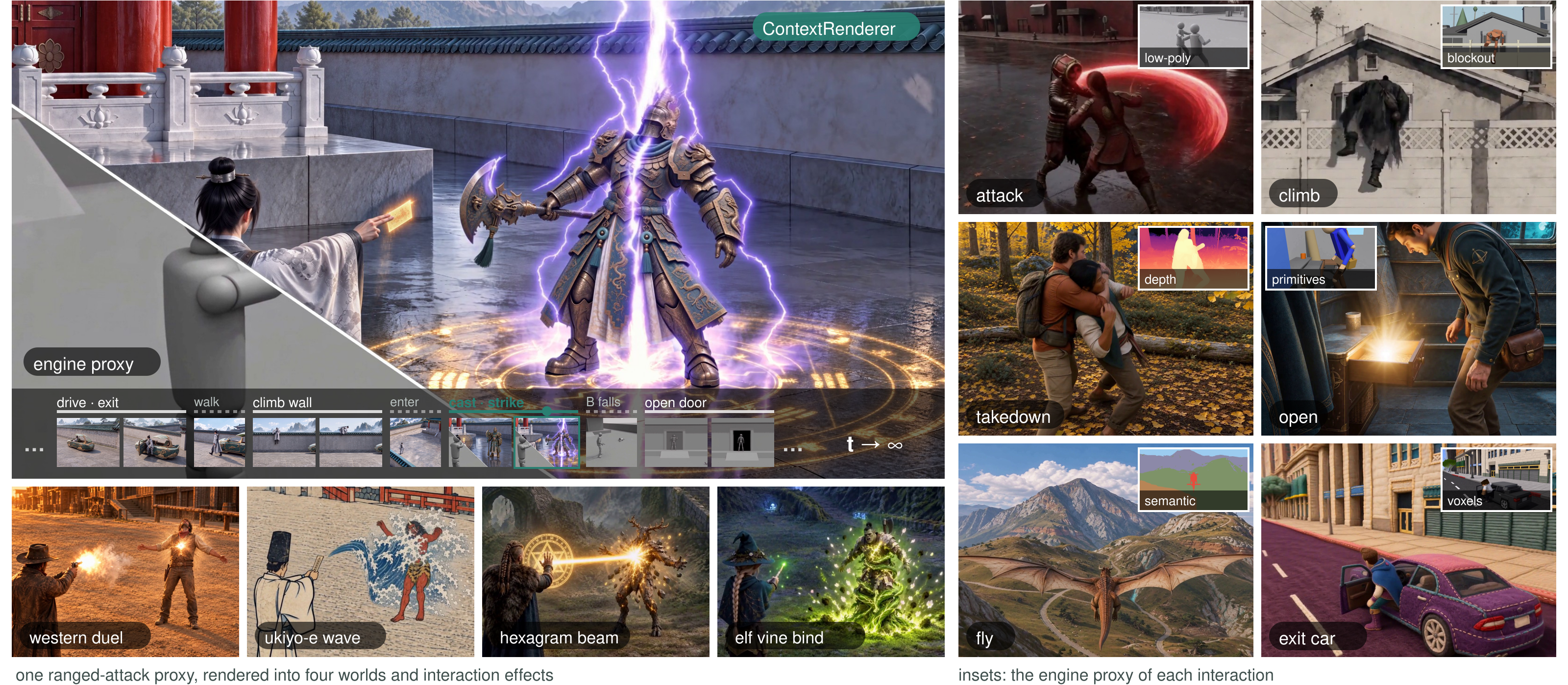}
\captionsetup{hypcap=false}
\captionof{figure}{\textbf{Rendering evolving game worlds across diverse proxies and appearances.} ContextRenderer generates successive video chunks from refreshed world context and visual history.}
\label{fig:1}
\end{minipage}
\par
}{}{\PackageError{paper}{Unable to attach teaser to title}{}}
\makeatother

\begin{document}
\maketitle

\begin{abstract}
Interactive game world models separate executable world evolution from visual generation, using video models as generative renderers to turn engine proxies into detailed observations. During sustained gameplay, however, existing renderers struggle to track evolving conditions without losing long-horizon continuity, and rarely generalize across diverse proxy formats and target styles. We therefore introduce \textbf{ContextRenderer}, an autoregressive video diffusion framework designed for both dynamic conditioning and cross-proxy generalization. It couples \textbf{Dynamic-Context Learning (DCL)} for chunk-level interaction supervision with \textbf{Recurrent History Compression (RHC)} to follow condition updates while preserving visual continuity within a bounded token budget. Across proxy representations, \textbf{Cross-Representation Flow Alignment (CRFA)} connects proxy reconstruction with appearance and interaction variation through a shared high-fidelity reconstruction. \textbf{WorldPrism}, our interaction-centric dataset constructed from agent-built worlds, AAA gameplay, and open-source videos, supplies the corresponding renderings and chunk-level annotations for these learning tasks. Experiments on GWM-Bench and Solar-Bench show improved reconstruction and visual quality, while qualitative results demonstrate minute-long interactive rendering under changing prompts and subject references.\ifdefined\ArxivMode\ \textbf{Project page:} {\urlstyle{same}\url{https://rainzor.github.io/contextrenderer/}}\fi
\end{abstract}

\section{Introduction}
\label{sec:intro}

Video generation models can synthesize diverse, temporally coherent scenes under flexible text and visual control~\cite{wan,minimaxh3,unividx}, and video world models extend these capabilities to action-responsive gaming environments~\cite{genie,gamengen,lingbotworld,actworld}. Yet visual quality alone does not guarantee accurate geometry, rule-consistent interactions, or persistent state~\cite{vbench,videophy2,persistentstate}. Recent game world models address this by separating state execution from visual generation~\cite{codeworldmodel,programmableworldmodel}: an engine or agent-written program maintains world state, enforces rules, and outputs scene layouts and motion, while a video model acts as a generative renderer to turn these outputs into detailed observations~\cite{agentgarten,magpie,wizard,alayarendererflash,solarrenderer}.

Existing generative renderers learn this mapping from paired proxy and target videos~\cite{epe,diffusionrenderer}, typically using text prompts~\cite{coarsetoreal} or a reference first frame~\cite{vace} to specify target appearance. Autoregressive variants extend this setup to interactive play by generating successive chunks from visual history, with recent systems supporting streaming proxies~\cite{magpie,alayarendererflash,streamrenderh3}, online text updates~\cite{edenlive,matrixgame40}, or memory references~\cite{matrixgame40}. However, two limitations remain in extended gameplay. First, \textbf{limited support for sustained interaction}: beyond an initial reference frame, current streaming interfaces mainly accommodate changing proxy videos or text prompts, leaving mid-session updates to subject references and appearance poorly supported over long horizons. Second, \textbf{limited generalization across representations}: despite progress on multimodal controls~\cite{cosmostransfer,solarrenderer}, interactive renderers are typically trained on one proxy format, limiting transfer across proxy representations and target styles.

We therefore aim to build a generative renderer for sustained exploration and interaction across diverse game worlds, which requires addressing two challenges. First, \textbf{dynamic conditioning for long-horizon interaction}: updated conditions may specify a state that differs from what is shown in visual history. The renderer must follow these updates while preserving unaffected content and temporal continuity~\cite{selfforcing,longlive}, and retain relevant history within bounded memory and computation~\cite{worldmem,framepack,helios,tinyhistory}. Second, \textbf{diverse conditioning across proxy representations}: proxies with different appearances and levels of detail may describe the same trajectory~\cite{coarsetoreal,solarrenderer}. From such inputs, the renderer must recover shared layout, motion, and event semantics, and combine them with target conditions to generate detailed appearance, character motion, and interaction effects without altering event timing or outcomes.

To address these challenges, we rethink generative rendering from a game renderer's perspective: at each step, the renderer processes the contextual information of the evolving world projected onto the current viewpoint. Guided by this insight, we introduce \textbf{ContextRenderer} (Fig.~\ref{fig:1}), an autoregressive video diffusion framework that generates successive chunks from visual history and a refreshed \emph{world context} combining the current proxy chunk, interaction prompt, and active subject references. \textbf{Dynamic-Context Learning (DCL)} pairs chunk-level interaction supervision with aligned token organization, training the model to depict updated conditions while preserving unaffected content. Meanwhile, \textbf{Recurrent History Compression (RHC)} maintains a bounded \emph{history context} by retaining a first-frame anchor alongside multiscale compression of past chunks~\cite{framepack,helios,tinyhistory}, sustaining long-horizon continuity within a fixed token budget.

For generalization across proxy representations, \textbf{Cross-Representation Flow Alignment (CRFA)} bridges coarse proxies and target variations through a shared high-fidelity reconstruction of each trajectory. Mapping different proxies to this shared video aligns their layout, motion, and event semantics, whereas mapping it to varied targets decouples appearance and interaction synthesis. Jointly training both auxiliary paths alongside direct proxy-to-target flow matching allows single-pass inference from the input proxy without generating the intermediate video.

Learning these capabilities requires trajectories that pair multiple proxy formats and target styles with evolving interaction states. We construct \textbf{WorldPrism} from agent-built worlds, AAA gameplay, and open-source scene and motion videos~\cite{dl3dv,magpie} to provide this supervision. Each trajectory links \(m\) proxy representations and \(n\) target variants through a shared high-fidelity reconstruction, while synchronized chunks pair active subject references with prompts describing scene appearance, interactions, and state changes.

Experiments on GWM-Bench and Solar-Bench~\cite{solarrenderer} show improved reconstruction fidelity and visual quality over size-matched autoregressive baselines. Qualitative results further demonstrate minute-long interactive rendering under changing prompts and subject references, with diverse appearances and interaction effects across proxy representations (Fig.~\ref{fig:1}).

Our contributions are threefold:
\begin{itemize}
\item We propose \textbf{ContextRenderer}, an autoregressive video diffusion framework combining \textbf{DCL} and \textbf{RHC} to follow changing world conditions while preserving visual continuity with bounded history.
\item We introduce \textbf{Cross-Representation Flow Alignment (CRFA)}, which connects diverse proxies with target appearance and interaction variation through a shared high-fidelity reconstruction, supporting direct inference.
\item We construct \textbf{WorldPrism}, an interaction-centric, multi-proxy dataset providing trajectory renderings, chunk-level interaction annotations, and subject references.
\end{itemize}

\section{Related Work}
\label{sec:related}

\noindent\textbf{Interactive world models.}
Interactive world models generate visual observations in response to user actions. Genie~\cite{genie} and GameNGen~\cite{gamengen} learn action-conditioned visual dynamics, while LingBot-World~\cite{lingbotworld} supports longer interactive rollouts. ActWorld~\cite{actworld} extends object interaction through chunk-level supervision and action-aware memory. Another line separates world evolution from visual generation: Code World Model~\cite{codeworldmodel} and Programmable World Model~\cite{programmableworldmodel} maintain world states and rules through executable programs, while video models render the resulting observations. Matrix-Game 4.0~\cite{matrixgame40} combines an executable game engine with an autoregressive renderer conditioned on low-poly proxies, dynamic captions, and memory references. Within this division, ContextRenderer focuses on learning continuous rendering under changing world conditions across diverse proxy representations while maintaining visual continuity.

\noindent\textbf{Generative rendering.}
Generative renderers condition on coarse renderings~\cite{coarsetoreal,magpie} or engine geometry and buffers~\cite{agentgarten,alayarendererflash}. StreamRender-H3~\cite{streamrenderh3} streams engine output with fixed session conditioning, while EdenLive~\cite{edenlive} supports dynamic prompt updates. SolarRenderer~\cite{solarrenderer} accepts diverse structural inputs and aligns structure tokens with the frames they describe, a design we adopt. These advances motivate a distinction beyond updating text or accepting multiple input formats. ContextRenderer organizes geometry, event descriptions, and subject references into a world context whose components can be replaced at chunk boundaries. It also imposes cross-rendering consistency on the generator's predictions: different coarse depictions of one trajectory should yield the same canonical rendering. This auxiliary constraint targets invariance to input style while allowing the main rendering task to produce different target appearances.

\noindent\textbf{Long-horizon video generation.}
Long-horizon video generation follows two complementary directions: autoregressive rollout and context representation. The former studies how to sustain generation, with Self Forcing~\cite{selfforcing} training on self-generated rollouts to reduce the training--inference gap. The latter studies how past information is retained and accessed. Retrieval methods such as WorldMem~\cite{worldmem} select relevant past observations. Compression methods reduce history tokens: FramePack~\cite{framepack} and Helios~\cite{helios} use multiscale representations, while TinyHistory~\cite{tinyhistory} learns an encoder through frame-query pretraining and video-continuation adaptation. Recurrent approaches instead accumulate context in hidden states; TTT-Video~\cite{tttvideo} uses neural networks as these states. ContextRenderer uses compressed visual history for continuity alongside engine-supplied geometry, event descriptions, and subject references updated at each chunk.

\section{Method}\label{sec:method}
Our goal is to support sustained exploration and interaction in game worlds through continuous video generation. As interaction changes the world, the video model must depict updated content while preserving the visual continuity of unaffected elements. ContextRenderer organizes this process as successive chunk predictions conditioned on the evolving world and visual history (Fig.~\ref{fig:3}).

\subsection{Problem Formulation}\label{sec:prelim}\label{sec:overview}\label{sec:world}

\begin{figure}[!htb]
\centering
\includegraphics[width=\columnwidth]{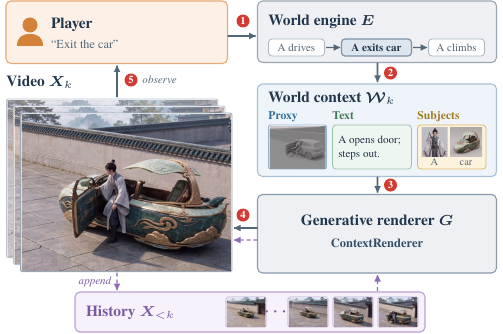}
\caption{\textbf{Interactive game world model.} Engine states provide current world conditions; the renderer combines them with visual history to generate the next observation. Outputs inform player actions and extend history. Images are illustrative.}
\label{fig:world-model}
\end{figure}

\begin{figure*}[t]
\centering
\includegraphics[width=\textwidth]{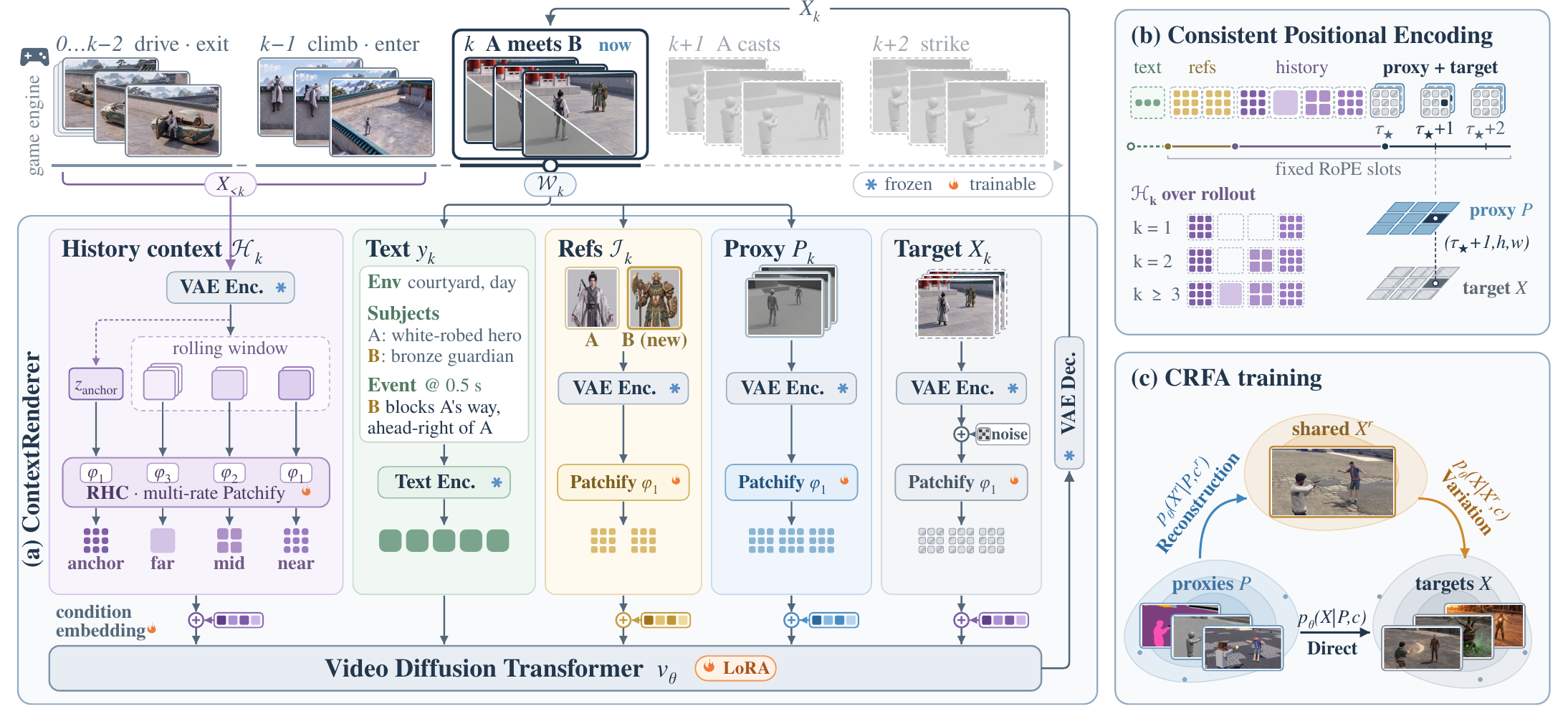}
\caption{\textbf{ContextRenderer overview.} (a) Refreshed text, subject references, and proxy chunks condition next-chunk generation alongside visual history. Recurrent History Compression keeps an anchor and multiscale rolling history; generated latents update future steps. (b) Proxy and target tokens share spatiotemporal RoPE coordinates at a fixed local temporal start after history; dashed text takes prefix RoPE slots (MiniMax-H3) or RoPE-free cross-attention (Wan). (c) Cross-Representation Flow Alignment jointly trains direct rendering, shared reconstruction, and reconstruction-to-target variation; inference uses the direct path. Scene images and token counts are schematic.}
\label{fig:3}
\label{fig:flow-alignment}
\end{figure*}

We consider interactive game world models in which players explore the world, interact with entities, and observe the consequences of their actions (Fig.~\ref{fig:world-model}). At time \(n\), a world engine \(E\) updates the world state \(s_n\) in response to player action \(a_n\), following executable rules~\cite{codeworldmodel,programmableworldmodel}. A generative renderer \(G\) then produces the next visual observation \(o_{n+1}\), using the updated state and preceding observations \(o_{\leq n}\):
\begin{equation}\label{eq:world-evolution}
s_{n+1}=E(s_n,a_n),
\qquad
o_{n+1}\sim G(\,\cdot\mid s_{n+1},o_{\leq n}).
\end{equation}

To render the evolving world with updated state, \(G\) needs a description of what is visible from the current viewpoint and how it should appear. We generate observations in video chunks. For each target chunk \(X_k\), the engine provides a proxy video chunk \(P_k\) covering the same time interval. This coarse representation conveys the visible scene layout, object motion, and interaction states to the video model, while leaving detailed appearance and interaction effects partially specified. A prompt \(y_k\) and subject reference images \(\mathcal I_k\) supply semantic and appearance requirements. Together, they form the world context \(\mathcal W_k=(P_k,y_k,\mathcal I_k)\), a view-dependent description refreshed as interaction unfolds.

The preceding video chunks form the visual history \(X_{<k}=(X_0,\ldots,X_{k-1})\), recording how the world has already appeared. At the chunk level, \(G\) therefore generates observations conditioned on the current world context and visual history: \(X_k\sim G(\,\cdot\mid\mathcal W_k,X_{<k})\). Each new chunk should reflect updated rendering requirements while preserving appearance and temporal continuity across chunks.

\subsection{ContextRenderer}\label{sec:training}

We implement \(G\) as an autoregressive video diffusion model built on pretrained models~\cite{wan,minimaxh3}. It learns the conditional distribution of the next video chunk given the current world context and visual history, denoted by \(p_\theta(X_k\mid\mathcal W_k,X_{<k})\), where \(\theta\) denotes the model parameters. This conditional generation process is learned through flow matching~\cite{flowmatching,rectifiedflow,sd3}, mapping Gaussian noise to the next chunk’s latent representation under these conditions. Dynamic-Context Learning (DCL) combines chunk-level interaction supervision with consistent local context organization, while Recurrent History Compression (RHC) constructs a bounded history context \(\mathcal H_k\) from \(X_{<k}\) for each prediction.

\noindent\textbf{Dynamic-Context Learning (DCL).}\label{sec:dcl}\label{sec:prediction}\label{sec:inference} During interaction, rendering conditions can change while visual history still depicts the preceding state. DCL trains the model to respond to these updates through chunk-level supervision. Each target chunk is paired with its corresponding proxy segment and a text prompt derived from its interaction annotations to describe the interactions within that interval. Subject references are updated when subjects or appearance requirements change and retained otherwise. Together with the history context from the same sequence, these pairs supervise the model to depict interaction-driven changes while preserving the appearance of unaffected content.

We fine-tune the pretrained Transformer as the conditional velocity network \(v_\theta\) through in-context learning. Clean visual context tokens are processed jointly with noisy target tokens, while the current prompt enters through the backbone’s text-conditioning pathway. For target latent \(z_0^{(k)}=\mathrm{Enc}(X_k)\), we construct \(z_t^{(k)}=(1-t)z_0^{(k)}+t\epsilon\), with \(\epsilon\sim\mathcal N(0,I)\) and \(t\in[0,1]\). Conditioned on \(\mathcal W_k\) and \(\mathcal H_k\), \(v_\theta\) is trained through conditional flow matching, with the loss evaluated only on the target chunk~\cite{flowmatching,rectifiedflow}:
\begin{equation}\label{eq:flow-matching}
\mathcal L_{\mathrm{FM}}
=
\mathbb E\!\left[
\left\|
v_\theta\!\left(z_t^{(k)},t\mid\mathcal W_k,\mathcal H_k\right)
-\left(\epsilon-z_0^{(k)}\right)
\right\|_2^2
\right].
\end{equation}
We maintain a consistent organization of token roles and local positions as content changes across chunks. Each visual token receives a learned role embedding:
\begin{equation}\label{eq:role-embedding}
\widetilde{\mathbf u}_i = \mathbf u_i+\mathbf e_{r(i)},
\end{equation}
where \(\mathbf u_i\) is the token feature and \(r(i)\) denotes its video, proxy, or reference-image role. History and target tokens share the video embedding, while all proxy representations share the proxy embedding. Flow-time embeddings distinguish clean visual conditions at \(t=0\) from the noisy target at the sampled time \(t\). Since proxy and target describe the same interval from the same viewpoint, corresponding tokens receive matching spatiotemporal RoPE coordinates:
\begin{equation}\label{eq:rope-alignment}
\rho_k^{P}(j,h,w)=\rho_k^{X}(j,h,w)=(\tau_\star+j,h,w).
\end{equation}
Here, \((j,h,w)\) indexes the local token grid, and \(\tau_\star\) fixes the target’s starting temporal position independently of the global chunk index \(k\). History occupies earlier positions, while reference images use separate slots.

\noindent\textbf{Recurrent History Compression (RHC).}\label{sec:rhc} As interaction continues, visual history grows beyond the budget of a single prediction. Recent observations supply detailed appearance and motion cues for continuation, while older observations provide broader visual context. Following multiscale history compression~\cite{helios}, RHC uses finer patches for recent frames and larger spatiotemporal patches for older ones. We encode each video chunk independently with the VAE and concatenate the resulting latent grids along time. A rolling window retains up to \(N\) recent latent frames, partitioned into recent, intermediate, and distant groups \(Z_k^{(1)},Z_k^{(2)},Z_k^{(3)}\). Each group is converted into a token sequence by $\phi_i$, a scale-specific 3D patch embedding followed by flattening, yielding the compressed history:
\begin{equation}\label{eq:history-context}
\mathcal H_k=
\big(
\phi_1(z_{\mathrm{anchor}})\ ;
\phi_3(Z_k^{(3)})\ ;
\phi_2(Z_k^{(2)})\ ;
\phi_1(Z_k^{(1)})
\big),
\end{equation}
where \(z_{\mathrm{anchor}}\) denotes the first-frame latent from the initial chunk \(X_0\), retained separately from the rolling window and embedded at the finest scale using \(\phi_1\). Fixed group capacities and patch sizes bound the total history token budget, including the anchor.

At inference, \(X_0\) is generated from \(\mathcal W_0\) without video history and initializes the rolling window. Subsequent chunks append latent frames, replacing the oldest entries once capacity is exceeded, and regroup retained frames for the next prediction under refreshed world conditions. Recent history occupies positions immediately before the fixed target position; unused earlier slots are zero-padded. Training covers empty, partially filled, and full windows under the same context organization, with backbone instantiations and anti-drifting training detailed in \supplref{Suppl.~A}{Suppl.~\ref{sec:suppl-arch}}.

\subsection{Cross-Representation Flow Alignment}\label{sec:flow-alignment}

Rendering from diverse coarse proxies requires both enriching their visual content and expressing it through varied appearances and interaction details. \textbf{Cross-Representation Flow Alignment (CRFA)} connects these learning tasks through a shared high-fidelity reconstruction \(X^r\). It enriches character modeling, scene appearance, and material detail while preserving the motion and event progression expressed by the proxies.

Omitting chunk indices, let \(P\) and \(X\) denote proxy and target chunks, and let \(c=(y,\mathcal I,\mathcal H)\) and \(c^r\) denote target- and reconstruction-side conditions. As illustrated in Fig.~\ref{fig:flow-alignment}(c), the model learns \(p_\theta(X^r\mid P,c^r)\) by reconstructing a shared high-fidelity video from different proxies of the same trajectory. Under the flow-matching objective, the conditional velocity fields are trained toward a common reconstruction, encouraging the model to recover shared layout, motion, and event semantics across proxy representations. The reconstruction then serves as a condition for learning \(p_\theta(X\mid X^r,c)\), enabling variations in appearance, environmental style, and interaction details while preserving the specified event timing and outcomes. We jointly learn these distributions alongside direct rendering \(p_\theta(X\mid P,c)\), using the same renderer and the conditional flow-matching objective in Sec.~\ref{sec:training}:
\begin{equation}\label{eq:total}
\mathcal L_{\mathrm{total}}
=\mathcal L_{P\rightarrow X}
+\lambda_1\mathcal L_{P\rightarrow X^r}
+\lambda_2\mathcal L_{X^r\rightarrow X}.
\end{equation}
The three terms supervise direct rendering, shared reconstruction, and reconstruction-conditioned target variation, respectively; we set \(\lambda_1=0.8\) and \(\lambda_2=0.2\). At inference, the renderer generates directly from the proxy and target-side conditions without producing \(X^r\).

\begin{figure}[t]
  \centering
  \includegraphics[width=\linewidth]{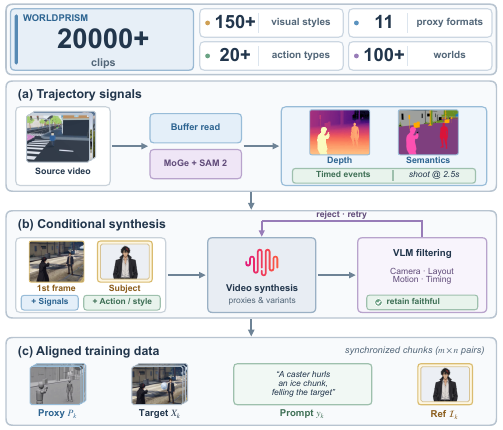}
  \caption{\textbf{Overview of the WorldPrism data engine.} (a) Structural and event signals are extracted from source trajectories. (b) Conditional synthesis and VLM filtering produce corresponding proxies and target variants. (c) Each trajectory's \(m\) proxies and \(n\) target variants form \(m\times n\) training pairs with synchronized chunks, interaction prompts, and subject references.}
  \label{fig:dataset}
\end{figure}

\begin{table*}[t]
\centering\small
\caption{\textbf{Reconstruction results on GWM-Bench and Solar-Bench.} We evaluate reconstruction fidelity (PSNR and LPIPS), motion and structural preservation (Motion Error and Edge F1), and temporal consistency (Adjacent-Frame Similarity, abbreviated as Adj. Sim.). All methods are autoregressive and grouped by model size. ContextRenderer achieves lower LPIPS in every displayed size-matched comparison. Best results within each benchmark are bolded.}
\label{tab:main-reconstruction}
\setlength{\tabcolsep}{3pt}
\begin{tabular*}{\textwidth}{@{\extracolsep{\fill}}cccccccc@{}}
\toprule
Dataset & Method & Params. & PSNR $\uparrow$ & LPIPS $\downarrow$ & Motion Error $\downarrow$ & Edge F1 $\uparrow$ & Adj. Sim. $\uparrow$ \\
\midrule
\multirow{8}{*}{\shortstack[c]{GWM-Bench-\\Reconstruction}} & EdenLive-5B-AR~\cite{edenlive} & 5B & 17.555 & 0.333 & 0.022 & 0.564 & 0.975 \\
 & ContextRenderer-5B (Ours) & 5B & 18.689 & 0.283 & 0.014 & 0.614 & 0.979 \\
 & EdenLive-14B-AR~\cite{edenlive} & 14B & 19.494 & 0.257 & 0.012 & 0.645 & 0.982 \\
 & ContextRenderer-14B (Ours) & 14B & 19.380 & 0.254 & 0.012 & \textbf{0.668} & 0.980 \\
 & Streaming-CWM~\cite{liao2026streamingcwm} & 33B & 16.613 & 0.377 & 0.044 & 0.482 & 0.982 \\
 & StreamRender-H3~\cite{streamrenderh3} & 33B & 15.416 & 0.470 & 0.035 & 0.411 & 0.982 \\
 & SolarRenderer-AR~\cite{solarrenderer} & 33B & 20.290 & 0.254 & 0.011 & 0.635 & \textbf{0.983} \\
 & ContextRenderer-33B (Ours) & 33B & \textbf{21.233} & \textbf{0.222} & \textbf{0.009} & 0.663 & \textbf{0.983} \\
\midrule
\multirow{2}{*}{\shortstack[c]{Solar-Bench-\\Reconstruction}} & SolarRenderer-AR~\cite{solarrenderer} & 33B & 18.660 & 0.318 & \textbf{0.149} & \textbf{0.256} & \textbf{0.967} \\
 & ContextRenderer-33B (Ours) & 33B & \textbf{19.010} & \textbf{0.298} & 0.172 & 0.226 & \textbf{0.967} \\
\bottomrule
\end{tabular*}
\end{table*}

\begin{figure*}[!t]
\centering
\includegraphics[width=\textwidth]{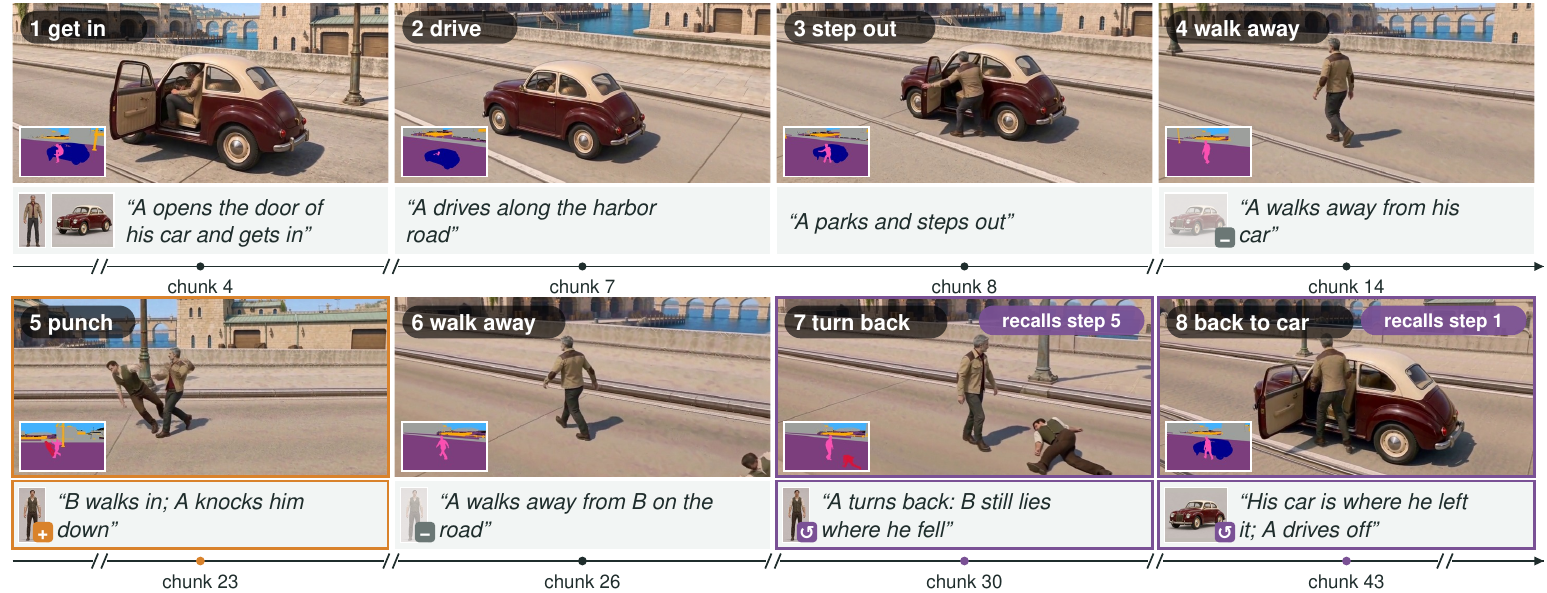}
\caption{\textbf{A minute-long interactive rendering session.} Prompts and subject references are updated across 48 generation chunks as gameplay unfolds, shown at representative chunks where later views recall the opponent and vehicle with consistent appearance. Symbols indicate references introduced (\textcolor[HTML]{D9822B}{+}), removed ($-$), or reintroduced (\textcolor[HTML]{7A5195}{$\circlearrowleft$}); insets show the semantic proxy, and // marks omitted intermediate chunks.}
\label{fig:qual-evolving}
\end{figure*}

\begin{table*}[t]
\centering\small
\caption{\textbf{Visual-variation results on GWM-Bench and Solar-Bench.} We evaluate visual quality using DOVER and VideoScore2, alongside motion and structural correspondence to reference videos and temporal consistency. All methods are autoregressive. Adj. Sim. denotes Adjacent-Frame Similarity; best reported results within each benchmark are bolded. N/A indicates unreliable DOVER and VideoScore2 estimates for Streaming-CWM outputs.}
\label{tab:main-variation}
\setlength{\tabcolsep}{3pt}
\begin{tabular*}{\textwidth}{@{\extracolsep{\fill}}cccccccc@{}}
\toprule
Dataset & Method & Params. & DOVER $\uparrow$ & VideoScore2 $\uparrow$ & Motion Error $\downarrow$ & Edge F1 $\uparrow$ & Adj. Sim. $\uparrow$ \\
\midrule
\multirow{8}{*}{\shortstack[c]{GWM-Bench-\\Variation}} & EdenLive-5B-AR~\cite{edenlive} & 5B & 0.444 & 3.067 & 0.054 & 0.445 & 0.971 \\
 & ContextRenderer-5B (Ours) & 5B & 0.469 & 3.096 & 0.055 & 0.448 & 0.975 \\
 & EdenLive-14B-AR~\cite{edenlive} & 14B & 0.460 & 3.141 & 0.051 & 0.461 & 0.978 \\
 & ContextRenderer-14B (Ours) & 14B & 0.522 & 3.250 & 0.046 & \textbf{0.469} & 0.982 \\
 & Streaming-CWM~\cite{liao2026streamingcwm} & 33B & N/A & N/A & 0.082 & 0.411 & 0.982 \\
 & StreamRender-H3~\cite{streamrenderh3} & 33B & 0.472 & 2.962 & 0.054 & 0.339 & 0.979 \\
 & SolarRenderer-AR~\cite{solarrenderer} & 33B & 0.477 & 3.133 & 0.055 & 0.417 & 0.975 \\
 & ContextRenderer-33B (Ours) & 33B & \textbf{0.527} & \textbf{3.318} & \textbf{0.040} & \textbf{0.469} & \textbf{0.983} \\
\midrule
\multirow{2}{*}{\shortstack[c]{Solar-Bench-\\Variation}} & SolarRenderer-AR~\cite{solarrenderer} & 33B & 0.557 & 4.712 & \textbf{0.060} & \textbf{0.294} & 0.975 \\
 & ContextRenderer-33B (Ours) & 33B & \textbf{0.583} & \textbf{4.854} & 0.066 & 0.212 & \textbf{0.980} \\
\bottomrule
\end{tabular*}
\end{table*}

\begin{figure*}[!t]
\centering
\includegraphics[width=\textwidth]{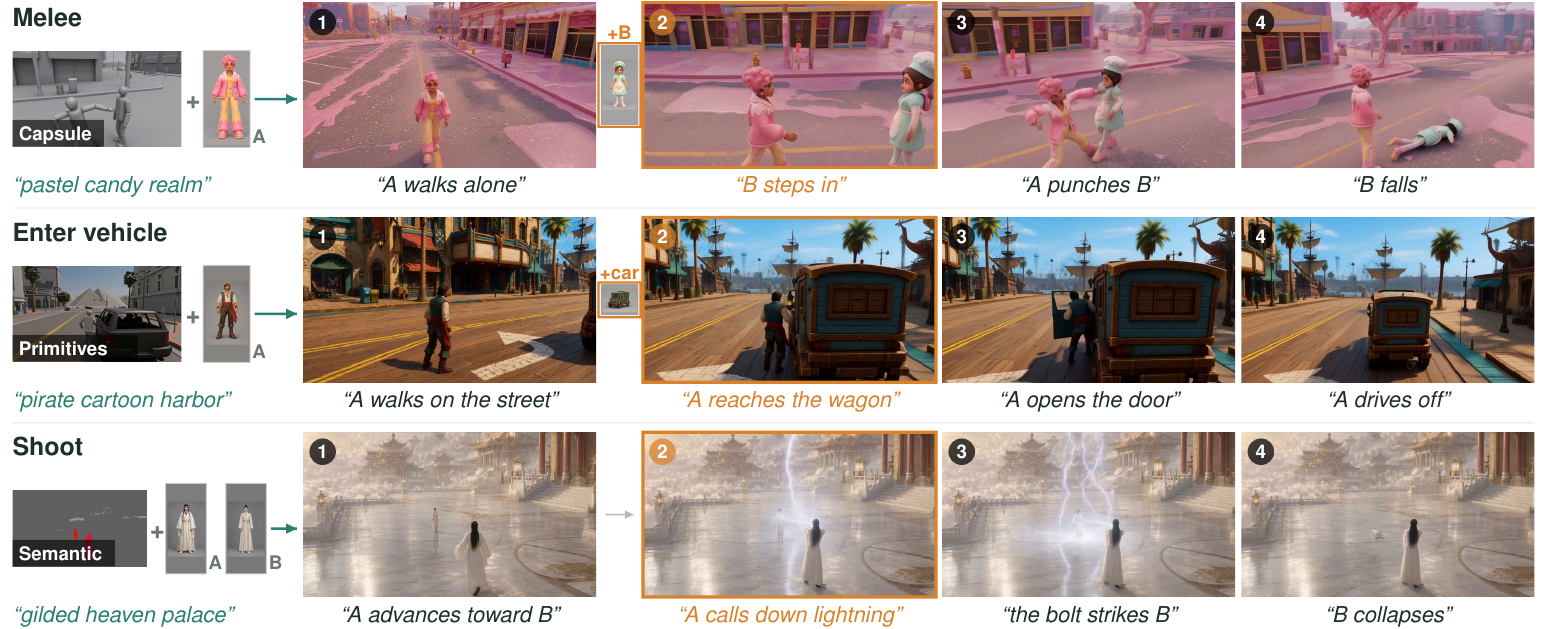}
\caption{\textbf{Any interaction, any proxy, any look.} Each row pairs one interaction with a different proxy representation, a subject reference, and a look prompt. Highlighted frames mark key events: a new subject reference (+B, +car) entering the world context, or a prompted effect (lightning).}
\label{fig:qual-any}
\end{figure*}

\subsection{WorldPrism Dataset}\label{sec:dataset}\label{sec:data-sources}\label{sec:data-construction}\label{sec:data-organization}\label{sec:data-annotation}\label{sec:data-analysis}

Learning across proxy representations and target variations requires corresponding videos that preserve the same underlying trajectory while varying its visual realization. We construct \textbf{WorldPrism}, an interaction-centric dataset providing these correspondences and chunk-level interaction annotations. Figure~\ref{fig:dataset} shows the construction pipeline.

\noindent\textbf{Paired video construction.} We collect source videos from agent-built worlds, AAA gameplay, driving simulations, and open-source scene datasets such as DL3DV~\cite{dl3dv}. Structural signals and interaction events are extracted from engines when available; otherwise, geometry, instance semantics, and event boundaries are estimated using MoGe~\cite{moge}, SAM 2~\cite{sam2}, and a vision-language model (VLM). Scene semantics and subject references guide the generation of a spatially aligned first frame, establishing the layout and subject appearance for subsequent video generation. Conditioned on this frame, structural videos, and action/style prompts, MiniMax-H3 Ref2VA~\cite{minimaxh3} generates coarse proxies and detailed target variants offline. The variants differ in character and scene appearance, motion details, and interaction effects. VLM filtering checks these requested changes against camera motion, spatial layout, and event timing to retain variants consistent with the source trajectory. For each trajectory, \(m\) proxy representations---including depth maps, semantic maps, low-poly renderings, and primitive renderings---are paired with \(n\) target variants to form \(m\times n\) pairs. A shared high-fidelity reconstruction \(X^r\), drawn from a source recording or generated as a canonical rendering, connects proxy reconstruction with target variation for CRFA.

\noindent\textbf{Interaction annotations and coverage.} We divide corresponding videos into temporally synchronized chunks and annotate each chunk with a prompt describing its scene appearance, active interactions, and state changes. Subject references are selected for the depicted subjects and updated when their appearance requirements change. Scene-only clips are annotated with environment descriptions and viewpoint motion. Together, these annotations provide chunk-level world conditions for DCL. WorldPrism contains 20K+ retained clips: 11K coarse proxy clips across 11 formats and 9K target variants across 150+ visual styles, covering 20+ interaction and locomotion categories (\supplref{Suppl.~B and~D}{Suppl.~\ref{sec:suppl-data} and~\ref{sec:suppl-prompt-templates}}).

\section{Experiments}
\label{sec:experiments}
We evaluate ContextRenderer on reconstruction, visual variation, and long-horizon interaction across diverse proxies and styles. We first compare with size-matched autoregressive renderers quantitatively and qualitatively, then ablate DCL, RHC, and CRFA under mid-clip updates.

\subsection{Experimental Setup}
\label{sec:protocol}\label{sec:implementation}
\noindent\textbf{Benchmarks and tasks.} We construct \textbf{GWM-Bench} with 200 reconstruction pairs with real targets and 120 visual-variation pairs with generated references for alternative appearances and interaction details under proxy-specified content. Solar-Bench uses Heldout53 for reconstruction and Hard208 for variation \cite{solarrenderer}. Clips contain 124 (GWM) or 158 (Solar) frames at 24 fps; results are aggregated separately by benchmark and task.

\noindent\textbf{Implementation and baselines.} ContextRenderer uses pretrained Wan (5B/14B) and MiniMax-H3 (33B)~\cite{wan,wan22,minimaxh3}. Wan generates 33-frame chunks (9 latent frames) at \(640\times384\) with 50 denoising steps; H3 generates 34-frame chunks (10 latent frames) at \(1344\times768\) with 30 steps. The final H3 chunk may be shorter. Our reference configuration uses 2,000 updates on 8 GPUs at learning rate \(2\times10^{-5}\). GWM baselines are EdenLive (5B/14B), SolarRenderer-AR, StreamRender-H3, and Streaming-CWM (33B)~\cite{edenlive,solarrenderer,streamrenderh3,liao2026streamingcwm}; the latter two use proxy video, target first frame, and a global prompt. Solar-Bench compares the 33B autoregressive models.

\noindent\textbf{Evaluation metrics.} PSNR and LPIPS measure reconstruction fidelity; DOVER and VideoScore2's visual-quality score assess visual variation \cite{lpips,dover,videoscore2}. Motion Error uses CoTracker3 trajectories, and Edge F1 measures structural adherence \cite{cotracker3,solarrenderer}. Adjacent-Frame Similarity uses consecutive-frame DINOv2 similarity to assess temporal consistency \cite{dinov2,solarrenderer}, with blinded VLM Elo in \supplref{Suppl.~C}{Suppl.~\ref{sec:suppl-exp}}.

\subsection{Reconstruction and Visual Variation}
\label{sec:reconstruction-variation}
\noindent\textbf{Reconstruction.} ContextRenderer improves perceptual reconstruction across scales (Table~\ref{tab:main-reconstruction}). The 33B model outperforms StreamRender-H3 and Streaming-CWM on all reported reconstruction metrics and matches SolarRenderer-AR in adjacent-frame similarity. Solar-Bench shows higher fidelity but weaker motion and structural preservation.

\noindent\textbf{Visual variation.} ContextRenderer improves visual quality and adjacent-frame similarity over size-matched EdenLive and SolarRenderer-AR (Table~\ref{tab:main-variation}). GWM quality, temporal consistency, and motion preservation improve with model size. Our 33B model exceeds StreamRender-H3 on all reported metrics and Streaming-CWM on motion, structure, and adjacent-frame similarity. SolarRenderer-AR better preserves motion and structure on Solar-Bench.

\subsection{Qualitative Results}
\noindent\textbf{Sustained interaction and long-horizon rendering.} Sustained interaction requires incorporating new rendering conditions without disrupting established content. Fig.~\ref{fig:qual-evolving} illustrates a minute-long session (48 chunks) across driving, exploration, and combat under successive prompt and subject-reference updates, where later views combine current engine-specified states with recognizable earlier appearances to maintain responsiveness and continuity.

\noindent\textbf{Diverse proxies and visual realizations.} Fig.~\ref{fig:qual-any} shows how ContextRenderer combines diverse proxy representations—capsule, primitive, and semantic—with prompts and dynamic subject references to generate varied character appearances, environments, and interaction effects. Presenting shooting as lightning illustrates variation in how an interaction is depicted while retaining event progression, alongside new subject references introduced mid-generation.

\begin{figure}[!t]
\centering
\includegraphics[width=\columnwidth]{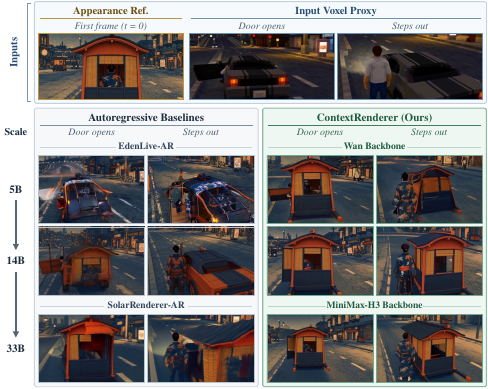}
\caption{\textbf{Visual comparison across model scales.} Vehicle exit from a voxel proxy and an ukiyo-e first frame, comparing ContextRenderer with EdenLive (5B/14B) and SolarRenderer-AR (33B).}
\label{fig:qual-comparison}
\end{figure}

\noindent\textbf{Visual quality comparison.} Fig.~\ref{fig:qual-comparison} compares vehicle-exit rendering from a voxel proxy under an ukiyo-e appearance condition. ContextRenderer better preserves the wooden palanquin's structure and more coherently depicts the character's door-opening and exit actions than size-matched baselines, which exhibit car-like geometry or structural deformation. Across 5B, 14B, and 33B models, larger variants further improve structural stability and interaction coherence.

\subsection{Ablation Studies}
\label{sec:ablation}
Table~\ref{tab:ablation} examines how each component contributes when new subjects and interactions appear mid-clip.
\textbf{(1) DCL.} Because the proxy still supplies layout and motion, freezing prompts and references barely affects edge structure, but leaves newly arriving characters and state changes unspecified, hurting target appearance match and visual quality.
\textbf{(2) RHC.} Keeping only the first frame still yields reasonable per-chunk quality, yet removing recent frames breaks continuity between consecutive chunks, causing the largest jump in temporal flicker and weaker structure.
\textbf{(3) CRFA.} Unlike the two inference-time ablations, omitting shared-reconstruction training degrades visual quality, alignment, and structure all at once, confirming that bridging diverse proxies through a common reconstruction is essential for geometry and appearance (\supplref{Suppl.~Fig.~17}{Suppl.~Fig.~\ref{fig:suppl-vis-ablation}}).

\begin{table}[!t]
\centering
\caption{\textbf{Ablations on the dynamic-update hard subset of GWM-Bench-Variation ($n=63$).} Sequences include mid-clip character arrivals. On the same 33B model, w/o DCL freezes prompts and references and w/o RHC keeps only the first frame at inference; w/o CRFA uses direct proxy-to-target training. Metrics: DOVER~\cite{dover}, DISTS~\cite{dists}, Edge F1~\cite{solarrenderer}, Flicker~\cite{solarrenderer}, and Adj. Sim.~\cite{solarrenderer,dinov2}.}
\label{tab:ablation}
\setlength{\tabcolsep}{1.4pt}
\renewcommand{\arraystretch}{1.10}
\footnotesize
\begin{tabular*}{\columnwidth}{@{\extracolsep{\fill}}lccccc@{}}
\toprule
Variant & \mbox{DOVER$\uparrow$} & \mbox{DISTS$\downarrow$} & \mbox{Edge F1$\uparrow$} & \mbox{Flicker$\downarrow$} & \mbox{Adj. Sim.$\uparrow$}\\
\midrule
w/o DCL & 0.552 & 0.154 & 0.473 & 0.938 & 0.985\\
w/o RHC & 0.558 & 0.150 & 0.462 & 1.255 & 0.982\\
w/o CRFA & 0.488 & 0.208 & 0.387 & 1.087 & 0.977\\
\midrule
Full model & \textbf{0.563} & \textbf{0.146} & \textbf{0.477} & \textbf{0.894} & \textbf{0.986}\\
\bottomrule
\end{tabular*}
\end{table}

\section{Conclusion and Limitations}
\label{sec:conclusion}
We presented ContextRenderer for continuous generative rendering of interactive game worlds. Dynamic-Context Learning (DCL) learns responses to updated world conditions, while Recurrent History Compression (RHC) maintains multiscale visual history within a bounded token budget. Cross-Representation Flow Alignment (CRFA) connects diverse proxies with appearance and interaction variation through a shared high-fidelity reconstruction. WorldPrism supplies corresponding renderings and chunk-level interaction annotations for training. Experiments on GWM-Bench and Solar-Bench demonstrate improved perceptual reconstruction and visual quality, alongside minute-long rendering with changing prompts and subject references.

\noindent\textbf{Limitations.} Despite these results, bringing ContextRenderer to real-time play will require further acceleration through few-step distillation, quantization, and runtime optimization. Over much longer sessions, maintaining stability will benefit from stronger memory management, while finer control across broader worlds and styles will rely on richer proxy conditioning and larger-scale data.

\clearpage

{\small
\bibliographystyle{ieeenat_fullname}
\bibliography{references}

\begin{thebibliography}{57}
\providecommand{\natexlab}[1]{#1}
\providecommand{\url}[1]{\texttt{#1}}
\expandafter\ifx\csname urlstyle\endcsname\relax
  \providecommand{\doi}[1]{doi: #1}\else
  \providecommand{\doi}{doi: \begingroup \urlstyle{rm}\Url}\fi

\bibitem[Bai et~al.(2025)Bai, Cai, Chen, Chen, Chen, Cheng, Deng, Ding, Gao,
  Ge, et~al.]{qwen3vl}
Shuai Bai, Yuxuan Cai, Ruizhe Chen, Keqin Chen, Xionghui Chen, Zesen Cheng,
  Lianghao Deng, Wei Ding, Chang Gao, Chunjiang Ge, et~al.
\newblock {Qwen3-VL} technical report.
\newblock \emph{arXiv preprint arXiv:2511.21631}, 2025.

\bibitem[Bansal et~al.(2025)Bansal, Peng, Bitton, Goldenberg,
  et~al.]{videophy2}
Hritik Bansal, Clark Peng, Yonatan Bitton, Roman Goldenberg, et~al.
\newblock {VideoPhy-2: A Challenging Action-Centric Physical Commonsense
  Evaluation in Video Generation}.
\newblock \emph{arXiv preprint arXiv:2503.06800}, 2025.

\bibitem[Bruce et~al.(2024)Bruce, Dennis, Edwards, Parker-Holder, Shi, Hughes,
  Lai, Mavalankar, Steigerwald, Apps, Aytar, Bechtle, Behbahani, Chan, Heess,
  Gonzalez, Osindero, Ozair, Reed, Zhang, Zolna, Clune, de~Freitas, Singh, and
  Rockt{\"a}schel]{genie}
Jake Bruce, Michael Dennis, Ashley Edwards, Jack Parker-Holder, Yuge Shi,
  Edward Hughes, Matthew Lai, Aditi Mavalankar, Richie Steigerwald, Chris Apps,
  Yusuf Aytar, Sarah Bechtle, Feryal Behbahani, Stephanie Chan, Nicolas Heess,
  Lucy Gonzalez, Simon Osindero, Sherjil Ozair, Scott Reed, Jingwei Zhang,
  Konrad Zolna, Jeff Clune, Nando de Freitas, Satinder Singh, and Tim
  Rockt{\"a}schel.
\newblock {Genie: Generative Interactive Environments}.
\newblock \emph{arXiv preprint arXiv:2402.15391}, 2024.

\bibitem[Chen et~al.(2024)Chen, Monso, Du, Simchowitz, Tedrake, and
  Sitzmann]{diffusionforcing}
Boyuan Chen, Diego~Marti Monso, Yilun Du, Max Simchowitz, Russ Tedrake, and
  Vincent Sitzmann.
\newblock Diffusion forcing: Next-token prediction meets full-sequence
  diffusion.
\newblock In \emph{Advances in Neural Information Processing Systems}, 2024.

\bibitem[Chen et~al.(2025)Chen, Jiang, Zheng, Chen, Su, and Zhu]{gft}
Huayu Chen, Kai Jiang, Kaiwen Zheng, Jianfei Chen, Hang Su, and Jun Zhu.
\newblock Visual generation without guidance.
\newblock \emph{arXiv preprint arXiv:2501.15420}, 2025.

\bibitem[Chen et~al.(2026{\natexlab{a}})Chen, Li, Kong, Zhu, Xu, Xiao, Guo, Ye,
  Zhang, Zhao, and Rao]{unividx}
Houyuan Chen, Hong Li, Xianghao Kong, Tianrui Zhu, Shaocong Xu, Weiqing Xiao,
  Yuwei Guo, Chongjie Ye, Lvmin Zhang, Hao Zhao, and Anyi Rao.
\newblock {UniVidX: A Unified Multimodal Framework for Versatile Video
  Generation via Diffusion Priors}.
\newblock \emph{ACM Transactions on Graphics}, 45\penalty0 (4),
  2026{\natexlab{a}}.

\bibitem[Chen et~al.(2026{\natexlab{b}})Chen, Lin, and Zhang]{codeworldmodel}
Yiwen Chen, Guosheng Lin, and Chi Zhang.
\newblock {Code World Model: Coding Agent as World Brain}.
\newblock \emph{arXiv preprint arXiv:2608.25927}, 2026{\natexlab{b}}.

\bibitem[Chi et~al.(2026)Chi, Miao, Shi, Wu, Wang, Chen, Dai, Li, Ren, Gao,
  Long, Duan, Lyu, Wu, and Liu]{agentgarten}
Jiawei Chi, Shangchen Miao, Zhiyuan Shi, Kailu Wu, Hanyang Wang, Weiliang Chen,
  Qiyu Dai, Chengyang Li, Jinshan Ren, Jun Gao, Mingsheng Long, Yueqi Duan,
  Jiangran Lyu, Jialong Wu, and Fangfu Liu.
\newblock {AgentGarten: Code Worlds for Evolving Agents}.
\newblock \emph{arXiv preprint arXiv:2610.12374}, 2026.

\bibitem[Chung et~al.(2023)Chung, Constant, Garcia, Roberts, Tay, Narang, and
  Firat]{umt5}
Hyung~Won Chung, Noah Constant, Xavier Garcia, Adam Roberts, Yi Tay, Sharan
  Narang, and Orhan Firat.
\newblock {UniMax}: Fairer and more effective language sampling for large-scale
  multilingual pretraining.
\newblock In \emph{International Conference on Learning Representations}, 2023.

\bibitem[Dalal et~al.(2025)Dalal, Koceja, Xu, Zhao, Han, Cheung, Kautz, Choi,
  Sun, and Wang]{tttvideo}
Karan Dalal, Daniel Koceja, Jiarui Xu, Yue Zhao, Shihao Han, Ka~Chun Cheung,
  Jan Kautz, Yejin Choi, Yu Sun, and Xiaolong Wang.
\newblock {One-Minute Video Generation with Test-Time Training}.
\newblock In \emph{Proceedings of the IEEE/CVF Conference on Computer Vision
  and Pattern Recognition}, pages 17702--17711, 2025.

\bibitem[Ding et~al.(2020)Ding, Ma, Wang, and Simoncelli]{dists}
Keyan Ding, Kede Ma, Shiqi Wang, and Eero~P. Simoncelli.
\newblock Image quality assessment: Unifying structure and texture similarity.
\newblock \emph{IEEE Transactions on Pattern Analysis and Machine
  Intelligence}, 2020.

\bibitem[Dosovitskiy et~al.(2017)Dosovitskiy, Ros, Codevilla, Lopez, and
  Koltun]{dosovitskiy2017carla}
Alexey Dosovitskiy, German Ros, Felipe Codevilla, Antonio Lopez, and Vladlen
  Koltun.
\newblock {CARLA}: An open urban driving simulator.
\newblock In \emph{Proceedings of the 1st Annual Conference on Robot Learning
  (CoRL)}, pages 1--16, 2017.

\bibitem[Esser et~al.(2024)Esser, Kulal, Blattmann, Entezari, M{\"u}ller,
  Saini, Levi, Lorenz, Sauer, Boesel, Podell, Dockhorn, English, Lacey,
  Goodwin, Marek, and Rombach]{sd3}
Patrick Esser, Sumith Kulal, Andreas Blattmann, Rahim Entezari, Jonas
  M{\"u}ller, Harry Saini, Yam Levi, Dominik Lorenz, Axel Sauer, Frederic
  Boesel, Dustin Podell, Tim Dockhorn, Zion English, Kyle Lacey, Alex Goodwin,
  Yannik Marek, and Robin Rombach.
\newblock {Scaling Rectified Flow Transformers for High-Resolution Image
  Synthesis}.
\newblock \emph{arXiv preprint arXiv:2403.03206}, 2024.

\bibitem[Fang et~al.(2026)Fang, Qian, Lu, Huang, Jiang, Wang, Song, Jiang, and
  Shou]{solarrenderer}
Guian Fang, Shengju Qian, Dongyue Lu, Junchao Huang, Lutao Jiang, Xin Wang,
  Yiren Song, Li Jiang, and Mike~Zheng Shou.
\newblock {SolarRenderer: Few-Step Generative Reference Rendering}.
\newblock \emph{arXiv preprint arXiv:2610.28630}, 2026.

\bibitem[Gomez-Nogales et~al.(2026)Gomez-Nogales, Hong, Ge, Zhuang,
  Comino-Trinidad, Casas, and Zhou]{coarsetoreal}
Gonzalo Gomez-Nogales, Yicong Hong, Chongjian Ge, Peiye Zhuang, Marc
  Comino-Trinidad, Dan Casas, and Yi Zhou.
\newblock {Coarse-to-Real: Generative Rendering for Populated Dynamic Scenes}.
\newblock \emph{arXiv preprint arXiv:2601.22301}, 2026.

\bibitem[He et~al.(2025)He, Jiang, Nie, et~al.]{videoscore2}
Xuan He, Dongfu Jiang, Ping Nie, et~al.
\newblock {VideoScore2}: Think before you score in generative video evaluation.
\newblock \emph{arXiv preprint arXiv:2509.22799}, 2025.

\bibitem[Ho and Salimans(2022)]{cfg}
Jonathan Ho and Tim Salimans.
\newblock Classifier-free diffusion guidance.
\newblock \emph{arXiv preprint arXiv:2207.12598}, 2022.

\bibitem[Huang et~al.(2025)Huang, Li, He, Zhou, and Shechtman]{selfforcing}
Xun Huang, Zhengqi Li, Guande He, Mingyuan Zhou, and Eli Shechtman.
\newblock {Self Forcing: Bridging the Train-Test Gap in Autoregressive Video
  Diffusion}.
\newblock \emph{arXiv preprint arXiv:2506.08009}, 2025.

\bibitem[Huang et~al.(2024)Huang, He, Yu, Zhang, et~al.]{vbench}
Ziqi Huang, Yinan He, Jiashuo Yu, Fan Zhang, et~al.
\newblock {VBench: Comprehensive Benchmark Suite for Video Generative Models}.
\newblock In \emph{Proceedings of the IEEE/CVF Conference on Computer Vision
  and Pattern Recognition}, 2024.

\bibitem[Huang et~al.(2026{\natexlab{a}})Huang, Xu, Peng, Zhang, Zheng, Song,
  Gu, Yu, Wang, and Cao]{edenlive}
Zehuan Huang, Tianxing Xu, Chunli Peng, Cheng Zhang, Zeyi Zheng, Yifei Song,
  Zekai Gu, Zhipeng Yu, Peng Wang, and Yanpei Cao.
\newblock {EdenLive: Real-Time Generative Rendering from Any Visual Reference}.
\newblock \url{https://vast-eden.ai/?p=edenlive}, 2026{\natexlab{a}}.
\newblock Accessed October 23, 2026.

\bibitem[Huang et~al.(2026{\natexlab{b}})Huang, Lin, Lin, Huang, Yu, Niu, Yang,
  Liu, Chuang, Zhang, and Wang]{programmableworldmodel}
Zheng-Hui Huang, Guixu Lin, Jiacheng Lin, Yi-Chuan Huang, Ruihan Yu, Muyao Niu,
  Siqi Yang, Yu-Lun Liu, Yung-Yu Chuang, Kaipeng Zhang, and Zhixiang Wang.
\newblock {Programmable World Model}.
\newblock \emph{arXiv preprint arXiv:2609.10540}, 2026{\natexlab{b}}.

\bibitem[Jiang et~al.(2025)Jiang, Han, Mao, Zhang, Pan, and Liu]{vace}
Zeyinzi Jiang, Zhen Han, Chaojie Mao, Jingfeng Zhang, Yulin Pan, and Yu Liu.
\newblock {VACE: All-in-One Video Creation and Editing}.
\newblock \emph{arXiv preprint arXiv:2503.07598}, 2025.

\bibitem[Karaev et~al.(2024)Karaev, Makarov, Wang, Neverova, Vedaldi, and
  Rupprecht]{cotracker3}
Nikita Karaev, Iurii Makarov, Jianyuan Wang, Natalia Neverova, Andrea Vedaldi,
  and Christian Rupprecht.
\newblock {CoTracker3}: Simpler and better point tracking by pseudo-labelling
  real videos.
\newblock \emph{arXiv preprint arXiv:2410.11831}, 2024.

\bibitem[Liang et~al.(2025)Liang, Gojcic, Ling, Munkberg, Hasselgren, Lin, Gao,
  Keller, Vijaykumar, Fidler, and Wang]{diffusionrenderer}
Ruofan Liang, Zan Gojcic, Huan Ling, Jacob Munkberg, Jon Hasselgren, Zhi-Hao
  Lin, Jun Gao, Alexander Keller, Nandita Vijaykumar, Sanja Fidler, and Zian
  Wang.
\newblock {DiffusionRenderer: Neural Inverse and Forward Rendering with Video
  Diffusion Models}.
\newblock \emph{arXiv preprint arXiv:2501.18590}, 2025.

\bibitem[Liao and Liu(2026)]{liao2026streamingcwm}
Tingting Liao and Guangyi Liu.
\newblock {Streaming Code World Model}.
\newblock \url{https://github.com/TingtingLiao/streaming-code-world-model},
  2026.

\bibitem[Lin et~al.(2026)Lin, Huang, Yang, Yang, Zhang, and
  Wang]{alayarendererflash}
Guixu Lin, Zheng-Hui Huang, Siqi Yang, Ming-Hsuan Yang, Kaipeng Zhang, and
  Zhixiang Wang.
\newblock {Generative World Renderer at the Speed of Play}.
\newblock \emph{arXiv preprint arXiv:2607.18703}, 2026.

\bibitem[Ling et~al.(2023)Ling, Sheng, Tu, et~al.]{dl3dv}
Lu Ling, Yichen Sheng, Zhi Tu, et~al.
\newblock {DL3DV-10K: A Large-Scale Scene Dataset for Deep Learning-based 3D
  Vision}.
\newblock \emph{arXiv preprint arXiv:2312.16256}, 2023.

\bibitem[Lipman et~al.(2022)Lipman, Chen, Ben-Hamu, Nickel, and
  Le]{flowmatching}
Yaron Lipman, Ricky T.~Q. Chen, Heli Ben-Hamu, Maximilian Nickel, and Matt Le.
\newblock {Flow Matching for Generative Modeling}.
\newblock \emph{arXiv preprint arXiv:2210.02747}, 2022.

\bibitem[Liu et~al.(2022)Liu, Gong, and Liu]{rectifiedflow}
Xingchao Liu, Chengyue Gong, and Qiang Liu.
\newblock {Flow Straight and Fast: Learning to Generate and Transfer Data with
  Rectified Flow}.
\newblock \emph{arXiv preprint arXiv:2209.03003}, 2022.

\bibitem[Lu et~al.(2026{\natexlab{a}})Lu, Zhu, Shi, Cai,
  et~al.]{persistentstate}
Jinpeng Lu, Dexu Zhu, Haoyuan Shi, Linghan Cai, et~al.
\newblock {Current World Models Lack a Persistent State Core}.
\newblock \emph{arXiv preprint arXiv:2606.20545}, 2026{\natexlab{a}}.

\bibitem[Lu et~al.(2026{\natexlab{b}})Lu, Ye, Li, Xue, Liu, Han, and
  Xie]{streamrenderh3}
Yanzuo Lu, Tian Ye, Yitong Li, Shuchen Xue, Haozhe Liu, Song Han, and Enze Xie.
\newblock {StreamRender-H3: Interactive Coding Worlds with Real-time Streaming
  Diffusion Render}.
\newblock \url{https://nvlabs.github.io/Sana/Sol-Engine/StreamRender-H3/},
  2026{\natexlab{b}}.
\newblock Accessed October 23, 2026.

\bibitem[{MiniMax}(2026)]{minimaxh3}
{MiniMax}.
\newblock {MiniMax H3}.
\newblock \url{https://github.com/MiniMax-AI/MiniMax-H3}, 2026.
\newblock Accessed October 23, 2026.

\bibitem[{NVIDIA} et~al.(2025){NVIDIA}, Alhaija, Alvarez, Bala, Cai,
  et~al.]{cosmostransfer}
{NVIDIA}, Hassan~Abu Alhaija, Jose Alvarez, Maciej Bala, Tiffany Cai, et~al.
\newblock {Cosmos-Transfer1: Conditional World Generation with Adaptive
  Multimodal Control}.
\newblock \emph{arXiv preprint arXiv:2503.14492}, 2025.

\bibitem[Oquab et~al.(2023)Oquab, Darcet, Moutakanni, et~al.]{dinov2}
Maxime Oquab, Timoth{\'e}e Darcet, Th{\'e}o Moutakanni, et~al.
\newblock {DINOv2}: Learning robust visual features without supervision.
\newblock \emph{arXiv preprint arXiv:2304.07193}, 2023.

\bibitem[Ravi et~al.(2024)Ravi, Gabeur, Hu, et~al.]{sam2}
Nikhila Ravi, Valentin Gabeur, Yuan-Ting Hu, et~al.
\newblock {SAM 2: Segment Anything in Images and Videos}.
\newblock \emph{arXiv preprint arXiv:2408.00714}, 2024.

\bibitem[Richter et~al.(2021)Richter, AlHaija, and Koltun]{epe}
Stephan~R. Richter, Hassan~Abu AlHaija, and Vladlen Koltun.
\newblock {Enhancing Photorealism Enhancement}.
\newblock \emph{arXiv preprint arXiv:2105.04619}, 2021.

\bibitem[{Robbyant Team} et~al.(2026){Robbyant Team}, Gao, Wang, Zeng, Zhu,
  Cheng, Li, Wang, Xu, Ma, Chen, Liu, Cheng, Yao, Zhu, Meng, Zheng, Bai, Chen,
  Shen, Yu, Zhu, Shen, and Ouyang]{lingbotworld}
{Robbyant Team}, Zelin Gao, Qiuyu Wang, Yanhong Zeng, Jiapeng Zhu, Ka~Leong
  Cheng, Yixuan Li, Hanlin Wang, Yinghao Xu, Shuailei Ma, Yihang Chen, Jie Liu,
  Yansong Cheng, Yao Yao, Jiayi Zhu, Yihao Meng, Kecheng Zheng, Qingyan Bai,
  Jingye Chen, Zehong Shen, Yue Yu, Xing Zhu, Yujun Shen, and Hao Ouyang.
\newblock {Advancing Open-source World Models}.
\newblock \emph{arXiv preprint arXiv:2601.20540}, 2026.

\bibitem[Su et~al.(2021)Su, Lu, Pan, Murtadha, Wen, and Liu]{rope}
Jianlin Su, Yu Lu, Shengfeng Pan, Ahmed Murtadha, Bo Wen, and Yunfeng Liu.
\newblock {RoFormer: Enhanced Transformer with Rotary Position Embedding}.
\newblock \emph{arXiv preprint arXiv:2104.09864}, 2021.

\bibitem[Valevski et~al.(2024)Valevski, Leviathan, Arar, and
  Fruchter]{gamengen}
Dani Valevski, Yaniv Leviathan, Moab Arar, and Shlomi Fruchter.
\newblock {Diffusion Models Are Real-Time Game Engines}.
\newblock \emph{arXiv preprint arXiv:2408.14837}, 2024.

\bibitem[{Wan Team}(2025)]{wan22}
{Wan Team}.
\newblock {Wan2.2}: Open and advanced large-scale video generative models.
\newblock \url{https://github.com/Wan-Video/Wan2.2}, 2025.

\bibitem[{Wan Team} et~al.(2025){Wan Team}, Wang, Ai, Wen, Mao, Xie, Chen, Yu,
  Zhao, Yang, Zeng, Wang, Zhang, Zhou, Wang, Chen, Zhu, Zhao, Yan, Huang, Feng,
  Zhang, Li, Wu, Chu, Feng, Zhang, Sun, Fang, Wang, Gui, Weng, Shen, Lin, Wang,
  Wang, Zhou, Wang, Shen, Yu, Shi, Huang, Xu, Kou, Lv, Li, Liu, Wang, Zhang,
  Huang, Li, Wu, Liu, Pan, Zheng, Hong, Shi, Feng, Jiang, Han, Wu, and
  Liu]{wan}
{Wan Team}, Ang Wang, Baole Ai, Bin Wen, Chaojie Mao, Chen-Wei Xie, Di Chen,
  Feiwu Yu, Haiming Zhao, Jianxiao Yang, Jianyuan Zeng, Jiayu Wang, Jingfeng
  Zhang, Jingren Zhou, Jinkai Wang, Jixuan Chen, Kai Zhu, Kang Zhao, Keyu Yan,
  Lianghua Huang, Mengyang Feng, Ningyi Zhang, Pandeng Li, Pingyu Wu, Ruihang
  Chu, Ruili Feng, Shiwei Zhang, Siyang Sun, Tao Fang, Tianxing Wang, Tianyi
  Gui, Tingyu Weng, Tong Shen, Wei Lin, Wei Wang, Wei Wang, Wenmeng Zhou, Wente
  Wang, Wenting Shen, Wenyuan Yu, Xianzhong Shi, Xiaoming Huang, Xin Xu, Yan
  Kou, Yangyu Lv, Yifei Li, Yijing Liu, Yiming Wang, Yingya Zhang, Yitong
  Huang, Yong Li, You Wu, Yu Liu, Yulin Pan, Yun Zheng, Yuntao Hong, Yupeng
  Shi, Yutong Feng, Zeyinzi Jiang, Zhen Han, Zhi-Fan Wu, and Ziyu Liu.
\newblock {Wan: Open and Advanced Large-Scale Video Generative Models}.
\newblock \emph{arXiv preprint arXiv:2503.20314}, 2025.

\bibitem[Wang et~al.(2025)Wang, Xu, Dai, Xiang, Deng, Tong, and Yang]{moge}
Ruicheng Wang, Sicheng Xu, Cassie Dai, Jianfeng Xiang, Yu Deng, Xin Tong, and
  Jiaolong Yang.
\newblock {MoGe: Unlocking Accurate Monocular Geometry Estimation for
  Open-Domain Images with Optimal Training Supervision}.
\newblock In \emph{Proceedings of the IEEE/CVF Conference on Computer Vision
  and Pattern Recognition}, pages 5261--5271, 2025.

\bibitem[Wu et~al.(2023)Wu, Zhang, Liao, Chen, Hou, Wang, Sun, Yan, and
  Lin]{dover}
Haoning Wu, Erli Zhang, Liang Liao, Chaofeng Chen, Jingwen Hou, Annan Wang,
  Wenxiu Sun, Qiong Yan, and Weisi Lin.
\newblock Exploring video quality assessment on user generated contents from
  aesthetic and technical perspectives.
\newblock In \emph{ICCV}, 2023.

\bibitem[Wu et~al.(2026)Wu, Lin, and Jin]{wizard}
Yuzhe Wu, Zijun Lin, and Yeying Jin.
\newblock {Wizard: Manipulate Any World in Real Time via Agentic Control}.
\newblock \emph{arXiv preprint arXiv:2610.40339}, 2026.

\bibitem[Xiao et~al.(2025)Xiao, Lan, Zhou, Ouyang, Yang, Zeng, and
  Pan]{worldmem}
Zeqi Xiao, Yushi Lan, Yifan Zhou, Wenqi Ouyang, Shuai Yang, Yanhong Zeng, and
  Xingang Pan.
\newblock {WorldMem: Long-term Consistent World Simulation with Memory}.
\newblock \emph{arXiv preprint arXiv:2504.12369}, 2025.

\bibitem[Xiong et~al.(2026)Xiong, Song, Kang, Yan, Jiang, Yang, Fu, Fotiadis,
  Wang, Liu, Liu, Yang, Lu, and Jacobs]{actworld}
Zhexiao Xiong, Yizhi Song, Hao Kang, Qing Yan, Liming Jiang, Jenson Yang,
  Zhoujie Fu, Stathi Fotiadis, Angtian Wang, Zichuan Liu, Bo Liu, Yiding Yang,
  Xin Lu, and Nathan Jacobs.
\newblock {ActWorld: From Explorable to Interactive World Model via
  Action-Aware Memory}.
\newblock \emph{arXiv preprint arXiv:2606.17730}, 2026.

\bibitem[Yang et~al.(2025)Yang, Huang, Chu, Xiao, Zhao, Wang, Li, Xie, Chen,
  Lu, Han, and Chen]{longlive}
Shuai Yang, Wei Huang, Ruihang Chu, Yicheng Xiao, Yuyang Zhao, Xianbang Wang,
  Muyang Li, Enze Xie, Yingcong Chen, Yao Lu, Song Han, and Yukang Chen.
\newblock {LongLive: Real-time Interactive Long Video Generation}.
\newblock \emph{arXiv preprint arXiv:2509.22622}, 2025.

\bibitem[Yin et~al.(2024{\natexlab{a}})Yin, Gharbi, Park, Zhang, Shechtman,
  Durand, and Freeman]{dmd2}
Tianwei Yin, Micha{\"e}l Gharbi, Taesung Park, Richard Zhang, Eli Shechtman,
  Fr{\'e}do Durand, and William~T. Freeman.
\newblock Improved distribution matching distillation for fast image synthesis.
\newblock In \emph{Advances in Neural Information Processing Systems},
  2024{\natexlab{a}}.

\bibitem[Yin et~al.(2024{\natexlab{b}})Yin, Gharbi, Zhang, Shechtman, Durand,
  Freeman, and Park]{dmd}
Tianwei Yin, Micha{\"e}l Gharbi, Richard Zhang, Eli Shechtman, Fr{\'e}do
  Durand, William~T. Freeman, and Taesung Park.
\newblock One-step diffusion with distribution matching distillation.
\newblock In \emph{Proceedings of the IEEE/CVF Conference on Computer Vision
  and Pattern Recognition}, 2024{\natexlab{b}}.

\bibitem[Yu et~al.(2026)Yu, Yuan, Yang, Qian, Wang, Wang, Yang, Liu, Li, Wang,
  and Ma]{dmad}
Zhengming Yu, Junkun Yuan, Haotian Yang, Gordon~Guocheng Qian, Yizhi Wang,
  Angtian Wang, Yiding Yang, Bo Liu, Xin Li, Wenping Wang, and Chongyang Ma.
\newblock {DMAD}: Distribution matching as adversarial distillation for fast
  visual generation.
\newblock \emph{arXiv preprint arXiv:2610.02188}, 2026.

\bibitem[Yuan et~al.(2026)Yuan, Yin, Li, Huang, Yang, and Yuan]{helios}
Shenghai Yuan, Yuanyang Yin, Zongjian Li, Xinwei Huang, Xiao Yang, and Li Yuan.
\newblock {Helios: Real Real-Time Long Video Generation Model}.
\newblock \emph{arXiv preprint arXiv:2603.04379}, 2026.

\bibitem[Zhan et~al.(2026)Zhan, Wang, Zhang, Zhu, Sun, Fang, Yu, Guo, and
  Fu]{magpie}
Xiaoyu Zhan, Xinyu Wang, Xiaohong Zhang, Huanjie Zhu, Tengjiao Sun, Pengcheng
  Fang, Jiaxing Yu, Yanwen Guo, and Dongjie Fu.
\newblock {Magpie: Real-Time World Renderer for Interactive Games}.
\newblock \emph{arXiv preprint arXiv:2608.27168}, 2026.

\bibitem[Zhang et~al.(2026{\natexlab{a}})Zhang, Qian, Kong, Jiang, Wang, Chen,
  Li, Chen, Wu, Chen, Li, Huang, Zhang, Li, Yu, Chen, Cheng, and
  Zhou]{matrixgame40}
Jihai Zhang, Runjia Qian, Kaiqin Kong, Biao Jiang, Puyi Wang, Siru Chen, Tianle
  Li, Guanjie Chen, Junyi Wu, Chubin Chen, Yuanhang Li, Kaichen Huang, Haobo
  Zhang, Yangguang Li, Wei Yu, Guibin Chen, Yu Cheng, and Yahui Zhou.
\newblock {Matrix-Game 4.0: Harnessing Video Models for Real-Time Playable Game
  Worlds}.
\newblock \emph{arXiv preprint arXiv:2610.27999}, 2026{\natexlab{a}}.

\bibitem[Zhang et~al.(2025)Zhang, Cai, Li, Wetzstein, and Agrawala]{framepack}
Lvmin Zhang, Shengqu Cai, Muyang Li, Gordon Wetzstein, and Maneesh Agrawala.
\newblock {Frame Context Packing and Drift Prevention in Next-Frame-Prediction
  Video Diffusion Models}.
\newblock \emph{arXiv preprint arXiv:2504.12626}, 2025.

\bibitem[Zhang et~al.(2026{\natexlab{b}})Zhang, Cai, Li, Zeng, Lu, Rao, Han,
  Wetzstein, and Agrawala]{tinyhistory}
Lvmin Zhang, Shengqu Cai, Muyang Li, Chong Zeng, Beijia Lu, Anyi Rao, Song Han,
  Gordon Wetzstein, and Maneesh Agrawala.
\newblock {TinyHistory: Lightweight Video History Embeddings via Two-Stage
  Context Learning}.
\newblock \emph{arXiv preprint arXiv:2512.23851}, 2026{\natexlab{b}}.

\bibitem[Zhang et~al.(2018)Zhang, Isola, Efros, Shechtman, and Wang]{lpips}
Richard Zhang, Phillip Isola, Alexei~A. Efros, Eli Shechtman, and Oliver Wang.
\newblock The unreasonable effectiveness of deep features as a perceptual
  metric.
\newblock In \emph{CVPR}, 2018.

\bibitem[Zhu et~al.(2026)Zhu, Zhao, He, Su, Li, and Zhu]{causalforcing}
Hongzhou Zhu, Min Zhao, Guande He, Hang Su, Chongxuan Li, and Jun Zhu.
\newblock Causal forcing: Autoregressive diffusion distillation done right for
  high-quality real-time interactive video generation.
\newblock \emph{arXiv preprint arXiv:2602.02214}, 2026.

\end{thebibliography}
}

\clearpage
\setcounter{section}{0}
\renewcommand{\thesection}{\Alph{section}}
\renewcommand{\theHsection}{\Alph{section}}
\setlength{\abovecaptionskip}{.45em}

\twocolumn[{%
  \begin{center}
    \includegraphics[width=\textwidth]{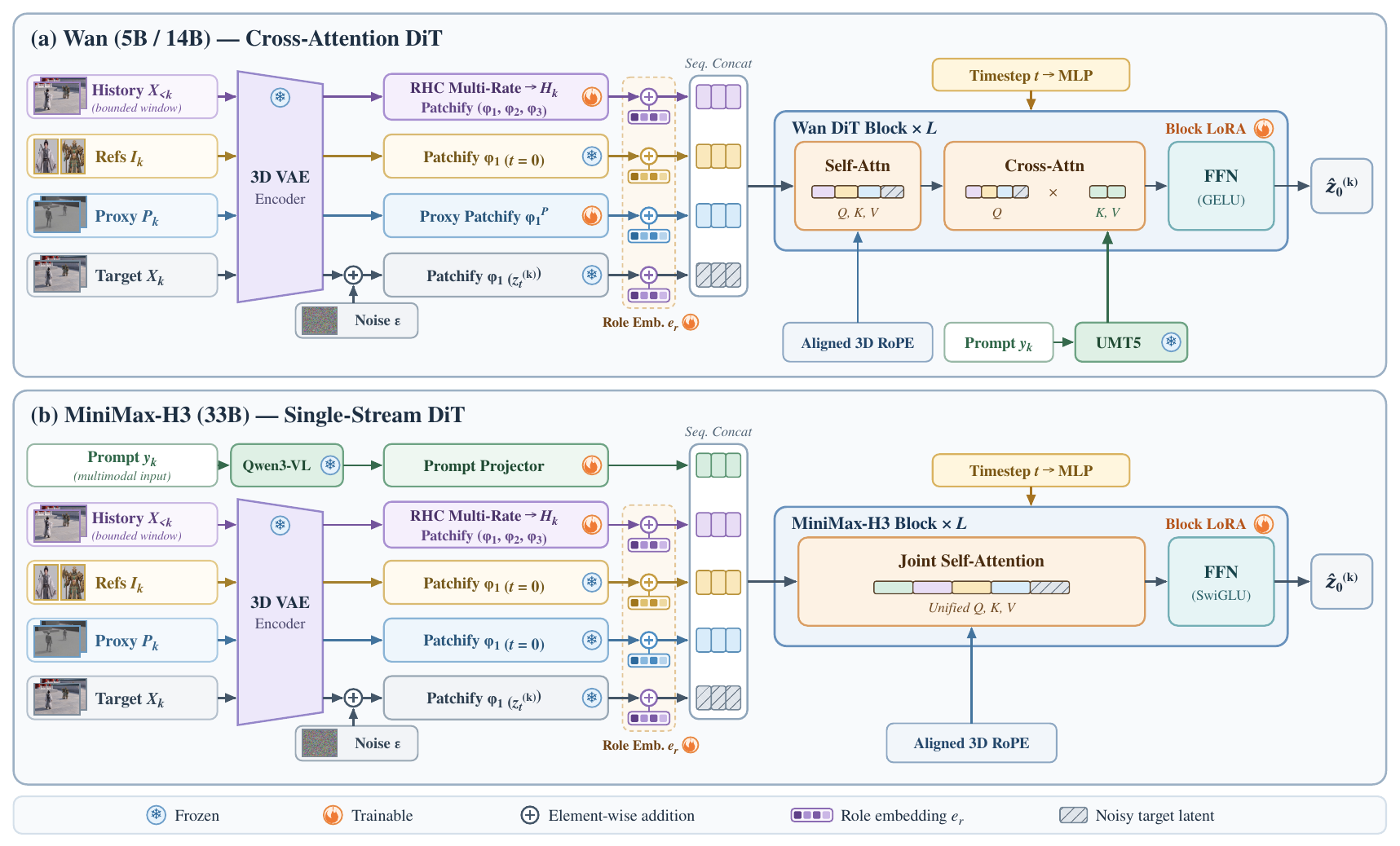}
  \end{center}
  \vspace{-0.6em}
  \captionof{figure}{\textbf{Dual-backbone instantiation of ContextRenderer.} Both video DiT families realize the unified interface $(\mathcal{W}_k, \mathcal{H}_k)$ with frozen VAE/text encoders, multiscale RHC patchification $(\phi_1,\phi_2,\phi_3)$, per-stream role embeddings $e_r$ on the four visual streams $(\mathcal{H}_k, \mathcal{I}_k, P_k, z_t^{(k)})$, aligned 3D RoPE, and lightweight Block LoRA adapters. \textbf{(a) Cross-Attention DiT (Wan~5B/14B):} visual tokens (with trainable proxy patchifier $\phi_1^P$) undergo self-attention while UMT5-encoded $y_k$ enters via cross-attention. \textbf{(b) Single-Stream DiT (MiniMax-H3~33B):} Qwen3-VL prompt tokens pass through a trainable projector (without $e_r$) to join all four visual streams in unified joint self-attention.}
  \label{fig:suppl-dual-backbone}
  \vspace{1.2em}
}]

\section{Architecture, Training, and Inference}
\label{sec:suppl-arch}

Following the formulation and notation of \mainref{Sec.~3 in the main paper}{Sec.~\ref{sec:method}}, this section provides the backbone instantiations, hierarchical dynamic prompting design, training and few-step distillation procedures, and streaming rollout specifications of ContextRenderer. Unless noted otherwise, all \mainref{main-paper evaluations}{evaluations in Sec.~\ref{sec:experiments}} correspond to our multi-step autoregressive models, while few-step distillation (Sec.~\ref{sec:suppl-training}) is investigated separately as a denoising-step reduction strategy in Sec.~\ref{sec:suppl-exp}.

\subsection{Model Architecture}
\label{sec:suppl-backbones}

\begin{figure*}[t]
\centering
\includegraphics[width=\textwidth]{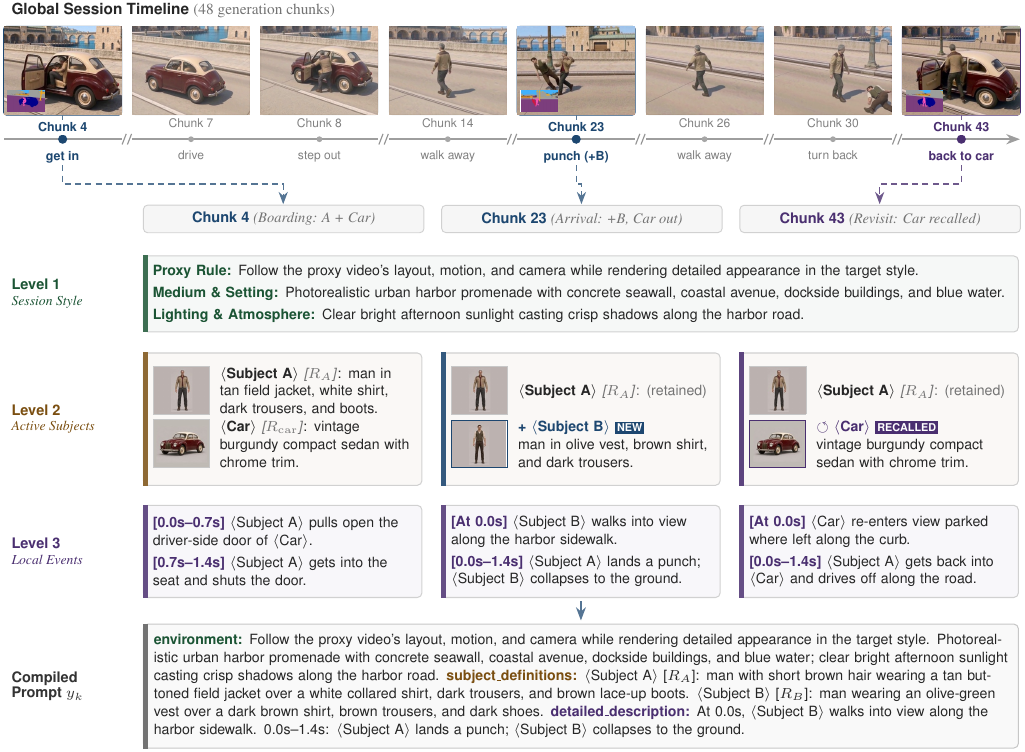}
\vspace{-0.3em}
\caption{\textbf{Event-driven three-level prompting over the minute-long interactive session (\mainref{main-paper Fig.~5}{Fig.~\ref{fig:qual-evolving}}).} Across the 48-chunk rollout in \mainref{main-paper Fig.~5}{Fig.~\ref{fig:qual-evolving}}, three representative chunks (Chunks~4, 23, and~43) are expanded with corresponding proxy insets in the lower-left corners. Level~1 fixes the global scene style and proxy-following rule; Level~2 dynamically admits, omits, and recalls active entities and their subject references ($R_A, R_{\mathrm{car}}, R_B$) by camera visibility; and Level~3 refreshes timestamped local events within each single generation chunk (\texttt{0.0s-1.4s}) without future leakage. The bottom panel illustrates the resulting compiled prompt $y_k$ at Chunk~23.}
\label{fig:suppl-prompt-timeline}
\end{figure*}

We validate ContextRenderer across distinct video diffusion backbones using a unified conditioning interface $(\mathcal{W}_k, \mathcal{H}_k)$, including DiTs with cross-attention text conditioning (Wan~5B/14B~\cite{wan,wan22}) and a single-stream DiT (MiniMax-H3~33B~\cite{minimaxh3}); see Fig.~\ref{fig:suppl-dual-backbone}. On Wan, the chunk-level conditioning prompt $y_k$---compiled from the session style, active subject descriptions, and chunk-local event annotations (Sec.~\ref{sec:suppl-conditioning})---is encoded by a frozen UMT5 text encoder~\cite{umt5} and injected via RoPE-free cross-attention, while the four visual streams ($\mathcal{H}_k, \mathcal{I}_k, P_k, z_t^{(k)}$) are concatenated into a unified visual self-attention sequence. On MiniMax-H3, the same chunk-level prompt $y_k$ (interleaved with active subject references and sampled proxy frames) is encoded by a frozen Qwen3-VL encoder~\cite{qwen3vl} and packed together with all four visual streams into a single joint self-attention sequence under prefix and 3D RoPE~\cite{rope}. Across both backbones, the multiscale RHC projections $(\phi_2, \phi_3)$ are mean-initialized from the native patch embedding $\phi_1$, while the clean frame-0 anchor $z_{\mathrm{anchor}}$ uses $\phi_1$ at stride $(1,2,2)$. The rolling window retains up to $N = 19$ latent frames on Wan with $(|Z_k^{(1)}|, |Z_k^{(2)}|, |Z_k^{(3)}|) = (1, 2, 16)$ at strides $(1,2,2)$, $(2,4,4)$, and $(4,8,8)$, and up to $N = 20$ latent frames ($4$ five-latent VAE groups) on MiniMax-H3 with $(5, 5, 10)$ at strides $(1,2,2)$, $(5,4,4)$, and $(5,8,8)$. Within each chunk, clean conditioning tokens $(\mathcal{W}_k, \mathcal{H}_k)$ do not attend to the noisy target $z_t^{(k)}$, whereas $z_t^{(k)}$ attends to all active conditioning streams; this asymmetric visibility keeps clean context representations invariant across denoising steps and enables exact intra-chunk KV reuse during single-stream inference.

\subsection{Hierarchical Dynamic Prompting}
\label{sec:suppl-conditioning}

In sustained interactive rendering, a static clip-level caption cannot track mid-rollout subject arrivals, revisits, or localized events without leaking future states. ContextRenderer instead structures each chunk's prompt and reference conditions into three functional levels refreshed along the rollout (Fig.~\ref{fig:suppl-prompt-timeline}). \textbf{Level~1 (Session Style)} defines the proxy-following rule together with global medium, setting, and lighting, remaining invariant across chunks unless a theme switch occurs. \textbf{Level~2 (Active Subjects)} maintains a visibility-gated registry that admits, omits, or recalls each entity's appearance description and subject reference according to whether it appears in the current camera view (determined by engine visibility online or per-chunk annotations offline). \textbf{Level~3 (Local Events)} specifies timestamped actions and state transitions strictly along the local chunk timeline, leaving spatial geometry and camera trajectory to the proxy $P_k$.

During long-horizon rollouts, ContextRenderer generates 34-frame chunks, spanning 48 chunks over the minute-long session in \mainref{main-paper Fig.~5}{Fig.~\ref{fig:qual-evolving}}, while Level~3 timestamps (\texttt{0.0s-1.4s}) specify events along each chunk's local timeline.

\subsection{Training and Distillation}
\label{sec:suppl-training}

We first train the multi-step autoregressive renderer on WorldPrism under CRFA using anti-drifting forcing and proxy-guided flow matching. To accelerate interactive inference, we further distill the converged renderer into a 4-step generator via decoupled DMD with frozen control weights.

\paragraph{Anti-drifting autoregressive forcing.}
Training a chunk-wise autoregressive renderer requires teaching the model to generate coherent continuations conditioned on visual history while resisting error accumulation~\cite{diffusionforcing,selfforcing} over long rollouts. Standard Teacher Forcing---conditioning each target chunk $z_t^{(k)}$ solely on clean ground-truth histories $\mathcal{H}_k$---optimizes efficiently, yet creates an exposure bias~\cite{selfforcing,helios} once $\mathcal{H}_k$ is populated by the model's own imperfect generations at inference time. Building on chunk-wise flow matching, we augment Teacher Forcing with two complementary anti-drifting strategies. First, inspired by Diffusion Forcing~\cite{diffusionforcing} and history corruption~\cite{helios}, we keep the global anchor $z_{\mathrm{anchor}}$ and current proxy $P_k$ strictly clean to preserve appearance grounding and crisp geometry, while perturbing historical target latents $z \in \bigcup_{m=1}^{3} Z_k^{(m)}$ with independent per-frame noise mixing:
\begin{equation}
\label{eq:suppl-hist-noise}
\tilde{z} = (1 - \rho) z + \rho \eta,\;\; \rho \sim \mathcal{U}(0, \rho_{\max}),\;\; \eta \sim \mathcal{N}(0, I),
\end{equation}
where we set $\rho_{\max} = 1/3$, paired with intermittent clean-history passes so the renderer learns to recover from imperfect context without sacrificing clean continuation fidelity. Second, following~\cite{helios}, we apply stochastic prefix truncation and history dropout during training, exposing a single model to empty, partially filled, and full RHC windows so it transitions smoothly from the opening chunk ($k = 0$) to steady-state rollouts without over-relying on historical inertia.

\paragraph{Proxy-guided flow matching.}
In autoregressive rendering, standard flow matching tends to over-rely on visual continuity from $\mathcal{H}_k$ and under-respond to abrupt structural updates in $P_k$. Rather than doubling per-step inference cost with classifier-free guidance (CFG)~\cite{cfg}, we follow Guidance-Free Training (GFT)~\cite{gft} and embed proxy guidance directly into training. Specifically, we pair the full condition $\mathcal{C}_k^+ = (\mathcal{W}_k, \mathcal{H}_k)$ with a weakened condition $\mathcal{C}_k^- = (\mathcal{W}_k^{-P}, \mathcal{H}_k)$ that drops only the proxy $P_k$ while retaining $(y_k, \mathcal{I}_k, z_{\mathrm{anchor}}, \mathcal{H}_k)$, with $10\%$ of samples training $\mathcal{C}_k^-$ directly under standard flow matching. Stopping gradients on the weak-condition pass of the same network $v_\theta$, we train at guidance scale $w \ge 1$:
\begin{equation}
\label{eq:suppl-pgfm}
\begin{aligned}
u_w^{(k)} &= v_\theta^- + w\big(u^{(k)} - v_\theta^-\big),\\
\mathcal{L}_{\mathrm{FM}}^{(w)}(\theta) &= \mathbb{E}\Big[\tfrac{1}{w^2}\big\|v_\theta^+ - \operatorname{sg}(u_w^{(k)})\big\|_2^2\Big],
\end{aligned}
\end{equation}
where $v_\theta^{\pm} = v_\theta(z_t^{(k)}, t \mid \mathcal{C}_k^{\pm})$ are the full- and weak-condition predictions of the same network $v_\theta$, $u^{(k)}$ is the flow-matching target velocity, and the $1/w^2$ scaling~\cite{gft} stabilizes training across guidance values. Setting $w = 1$ reduces Eq.~\eqref{eq:suppl-pgfm} to standard flow matching, while $w > 1$ (we use $w = 1.5$) strengthens structural adherence to $P_k$ in a single CFG-free forward pass; because $\mathcal{H}_k$ and $\mathcal{I}_k$ are retained in $\mathcal{C}_k^-$, the guidance focuses on proxy geometry rather than color or style.

\begin{algorithm}[!t]
\caption{Streaming Chunk Rollout}
\label{alg:suppl-rollout}
\scriptsize
\begin{algorithmic}[1]
\Require Engine $E$, initial state $s_0$, renderer $v_\theta$, 3D VAE decoder $\mathrm{Dec}$, schedule $\{t_S,\dots,t_0\}$.
\Ensure Rendered video stream $\{X_0, X_1, \dots\}$.
\State $\mathcal{M}_0 \leftarrow \varnothing,\;\; z_{\mathrm{anchor}} \leftarrow \varnothing$
\For{$k = 0, 1, 2, \dots$}
  \Statex \hspace{\algorithmicindent}\textcolor{black!50}{\textit{// Dynamic world context (Sec.~\ref{sec:suppl-conditioning})}}
  \State $(s_{k+1}, P_k) \leftarrow E(s_k, a_k)$
  \State $\mathcal{I}_k \leftarrow \textsc{SelectVisibleRefs}(s_{k+1})$
  \State $y_k \leftarrow \textsc{CompilePrompt}(s_{k+1}, \mathcal{I}_k, P_k)$
  \State $\mathcal{W}_k \leftarrow (P_k, y_k, \mathcal{I}_k)$
  \Statex \hspace{\algorithmicindent}\textcolor{black!50}{\textit{// Bounded memory retrieval}}
  \If{$k = 0$}
    \State $\mathcal{H}_0 \leftarrow \varnothing$
  \Else
    \State $(Z_k^{(1)}, Z_k^{(2)}, Z_k^{(3)}) \leftarrow \textsc{Retrieve}(\mathcal{M}_k)$
    \State $\mathcal{H}_k \leftarrow \textsc{RHC}(z_{\mathrm{anchor}}, Z_k^{(3)}, Z_k^{(2)}, Z_k^{(1)})$
  \EndIf
  \Statex \hspace{\algorithmicindent}\textcolor{black!50}{\textit{// Chunk-local flow integration}}
  \State $z_{t_S}^{(k)} \sim \mathcal{N}(0, I)$
  \For{$j = S, \dots, 1$}
    \State $v_j \leftarrow v_\theta(z_{t_j}^{(k)}, t_j \mid \mathcal{W}_k, \mathcal{H}_k)$
    \State $z_{t_{j-1}}^{(k)} \leftarrow \textsc{SolverStep}(z_{t_j}^{(k)}, v_j, t_j \rightarrow t_{j-1})$
  \EndFor
  \Statex \hspace{\algorithmicindent}\textcolor{black!50}{\textit{// Latent-first memory store \& decode}}
  \State $\hat{z}_0^{(k)} \leftarrow z_{t_0}^{(k)}$;\;\; \textbf{if} $k = 0$ \textbf{then} $z_{\mathrm{anchor}} \leftarrow \hat{z}_0^{(0)}[0]$
  \State $\mathcal{M}_{k+1} \leftarrow \textsc{Store}(\mathcal{M}_k, \hat{z}_0^{(k)})$
  \State $X_k \leftarrow \mathrm{Dec}(\hat{z}_0^{(k)})$
\EndFor
\end{algorithmic}
\end{algorithm}

\paragraph{Decoupled few-step distillation.}
Directly distilling all weights into a few-step generator can degrade learned proxy control and appearance fidelity. Following~\cite{solarrenderer}, we instead separate step compression from control adaptation to distill the multi-step renderer into a $4$-step generator. Specifically, we freeze $W_0$, the converged multi-step control adapter $\Delta W_{\mathrm{ctrl}}$, and all RHC projections $\{\phi_i\}$, and train only an acceleration adapter $\Delta W_{\mathrm{acc}}$ initialized from a pretrained few-step prior~\cite{dmad}:
\begin{equation}
\label{eq:suppl-dual-lora}
W_{\mathrm{student}} = \underbrace{W_0 + \Delta W_{\mathrm{ctrl}}}_{\text{frozen multi-step renderer}} + \Delta W_{\mathrm{acc}}.
\end{equation}
At deployment, $\Delta W_{\mathrm{acc}}$ is merged directly into the backbone weights without altering model topology. Denoting the trainable parameters of $\Delta W_{\mathrm{acc}}$ by $\psi$, we optimize $\psi$ via Distribution Matching Distillation (DMD)~\cite{dmd,dmd2,selfforcing}:
\begin{equation}
\label{eq:suppl-dmd-loss}
\begin{aligned}
\nabla_\psi \mathcal{L}_{\mathrm{DMD}} &= \mathbb{E}_{t,\epsilon}\Bigl[w(t)\bigl(f_{\mathrm{fake}}(\tilde{z}_t^{(k)}, t \mid \mathcal{C}_k) \\
&\quad - f_{\mathrm{real}}(\tilde{z}_t^{(k)}, t \mid \mathcal{C}_k)\bigr)\frac{\partial \hat{z}_0^{(k)}}{\partial \psi}\Bigr],
\end{aligned}
\end{equation}
where $\hat{z}_0^{(k)} = G_\psi(\mathcal{C}_k)$ is the 4-step student output, $\tilde{z}_t^{(k)}$ is its re-noised latent, $f_{\mathrm{real}}$ and $f_{\mathrm{fake}}$ are the frozen multi-step teacher and online-updated fake score denoisers, and $w(t)$ follows standard DMD2 weighting~\cite{dmd2}. Following~\cite{selfforcing,causalforcing}, distillation is performed over autoregressive rollouts using single-chunk backpropagation to bound memory.

\subsection{Streaming Inference}
\label{sec:suppl-infer}

Algorithm~\ref{alg:suppl-rollout} formalizes the closed-loop streaming rollout of \mainref{Fig.~2 in the main paper}{Fig.~\ref{fig:world-model}} across both backbone families. At each chunk step $k$, the world engine advances the scene from player action $a_k$ and refreshes the chunk context $\mathcal{W}_k = (P_k, y_k, \mathcal{I}_k)$ (Sec.~\ref{sec:suppl-conditioning}), after which ContextRenderer retrieves compressed history $\mathcal{H}_k$ from visual memory $\mathcal{M}_k$ and integrates $v_\theta$ from $t_S = 1$ to $t_0 = 0$ under local 3D RoPE coordinates via either multi-step ODE or distilled four-step solvers before decoding and streaming $X_k$ to the player. Across chunks, \textsc{Retrieve} and \textsc{Store} operate over the first-frame anchor $z_{\mathrm{anchor}}$ and a bounded latent buffer, keeping per-chunk compute and memory constant over arbitrary rollout horizons.

\section{WorldPrism Dataset and GWM-Bench}
\label{sec:suppl-data}

\begin{figure*}[!t]
  \centering
  \includegraphics[width=\textwidth]{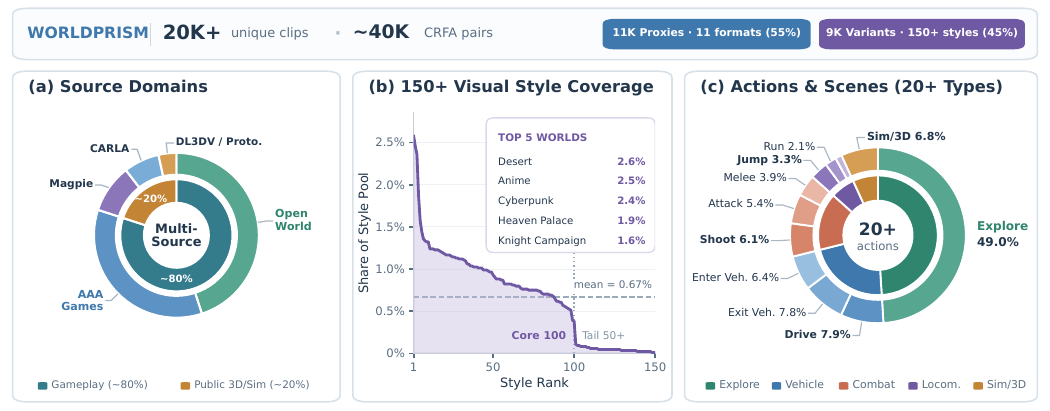}
  \caption{\textbf{Quantitative coverage distributions of WorldPrism.}
  \emph{Top:} Corpus scale aligned with \mainref{Sec.~3.4 of the main paper}{Sec.~\ref{sec:dataset}} ($20\text{K}+$ retained unique clips: $11\text{K}$ coarse proxies across $11$ formats and $9\text{K}$ target variants across $150+$ visual styles, expanding to $\sim 40\text{K}$ deduplicated CRFA pairs).
  \textbf{(a)} Source domain split between gameplay/open-world trajectories ($\sim 80\%$) and public 3D, simulation, and prototype sources ($\sim 20\%$: Magpie, CARLA, DL3DV, and 3D game prototypes).
  \textbf{(b)} Ranked share across $150+$ target visual worlds ($100$ core worlds and $50+$ long-tail expansion worlds).
  \textbf{(c)} Two-ring breakdown across $20+$ interaction, locomotion, and scene categories.}
  \label{fig:suppl-stats}
  \vspace{\dblfloatsep}
  \includegraphics[width=\textwidth]{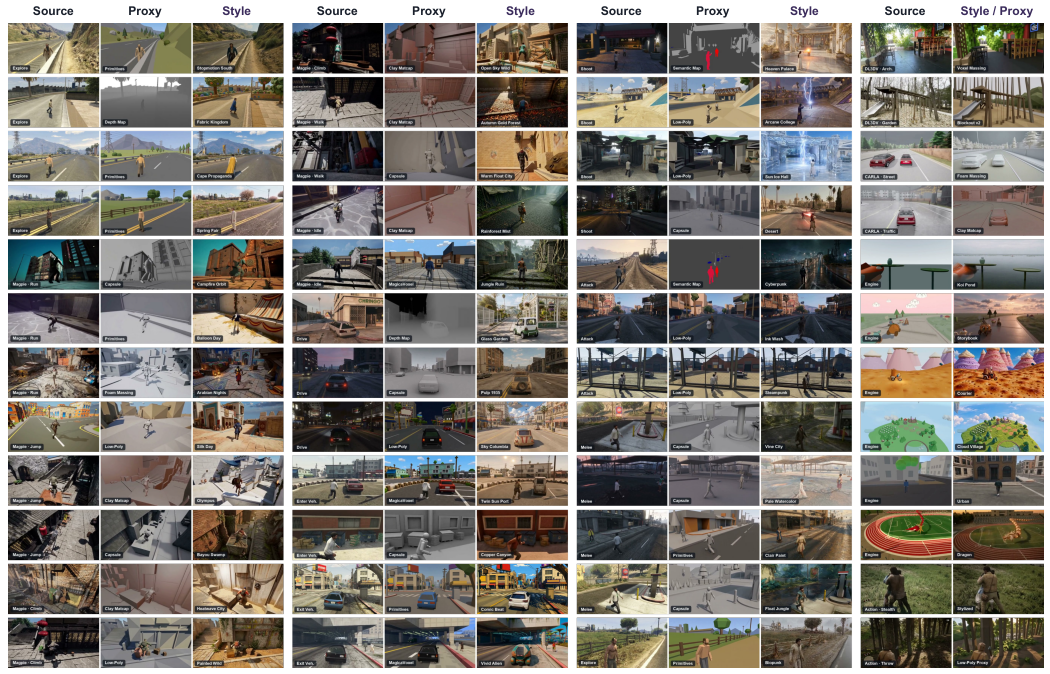}
  \caption{\textbf{Multi-source qualitative gallery of WorldPrism ($48$ distinct trajectories).}
  \emph{Left three blocks ($36$ triplets):} Open-world exploration, Magpie locomotion, and AAA gameplay paired with diverse coarse proxies (depth, semantic, voxel, primitive, capsule, low-poly, clay matcap, flat-toon, and massing) and stylized target realizations across $13$ action categories.
  \emph{Right block ($12$ pairs):} DL3DV real-world scenes and CARLA urban driving mapped to coarse proxies, alongside lightweight 3D game prototypes and action gameplay pairs.}
  \label{fig:suppl-gallery}
\end{figure*}

\subsection{WorldPrism Distribution and \texorpdfstring{$m\times n$}{m x n} Pairing}
\label{sec:suppl-worldprism}

\noindent\textbf{Distributional coverage.}
Complementing the pipeline overview in \mainref{Sec.~3.4 of the main paper}{Sec.~\ref{sec:dataset}}, Figs.~\ref{fig:suppl-stats} and~\ref{fig:suppl-gallery} summarize WorldPrism's source, style, and interaction coverage.
AAA gameplay and open-world exploration trajectories ($\sim 80\%$) supply sustained camera traversal and player-driven interactions---including vehicle boarding, driving, exiting, melee, shooting, stealth, and throwing---while public 3D and simulation corpora ($\sim 20\%$)---Magpie~\cite{magpie} agile locomotion, CARLA~\cite{dosovitskiy2017carla} urban driving, DL3DV~\cite{dl3dv} real-world scenes, and lightweight agent-built 3D prototypes---broaden geometric and environmental diversity.
Across the $150+$ target visual worlds (Fig.~\ref{fig:suppl-stats}b), $100$ core worlds provide dense multi-scene supervision and $50+$ long-tail worlds expand stylistic coverage across $20+$ action and scene categories (Fig.~\ref{fig:suppl-stats}c).

\noindent\textbf{$m\times n$ pairing and sampling.}
The $20\text{K}+$ clips reported in \mainref{Sec.~3.4}{Sec.~\ref{sec:dataset}} ($11\text{K}$ proxies + $9\text{K}$ target variants) count \emph{unique stored clips}, whereas CRFA training pairs are constructed strictly within each source trajectory $\tau$ (never across trajectories).
Given $m_\tau$ retained proxies, $n_\tau$ target variants, and (when available) one canonical high-fidelity reconstruction $X^r$, within-trajectory pairing induces
\begin{equation}
\label{eq:suppl-mn-pairs}
N_{\mathrm{pairs}}(\tau) = \underbrace{m_\tau n_\tau}_{P\rightarrow X} + \underbrace{m_\tau}_{P\rightarrow X^r} + \underbrace{n_\tau}_{X^r\rightarrow X}
\end{equation}
candidate pairs (or $m_\tau n_\tau$ when $X^r$ is absent).
For example, $m=5, n=4$ with one $X^r$ induces up to $29$ candidate pairs from $10$ clips.
Rather than exhaustively training on all possible proxy--target combinations, we sample from this within-trajectory candidate pool to balance the relative ratio across the three CRFA objectives, yielding \textbf{$\sim 40\text{K}$ deduplicated training pairs} ($\sim 19\text{K}$ $P\rightarrow X$, $\sim 17\text{K}$ $P\rightarrow X^r$, and $\sim 3.5\text{K}$ $X^r\rightarrow X$).
Train/test splitting is performed strictly at the \textbf{source-trajectory level prior to $m\times n$ expansion}, preventing any proxy, variant, or $X^r$ leakage.

\subsection{Prompt Design and Quality Control}
\label{sec:suppl-prompt-qc}

Following the structured conditioning conventions of modern generative renderers~\cite{solarrenderer}, WorldPrism separates \emph{offline synthesis prompts} from \emph{chunk-level training prompts} (full templates in Sec.~\ref{sec:suppl-prompt-templates}):

\noindent\textbf{Data synthesis and VLM verification (Fig.~\ref{fig:suppl-prompts}).}
Offline target and proxy synthesis uses a structured schema that explicitly decouples \emph{structural authority} (camera path, spatial slots, scale, and causal event timing from the control video) from \emph{visual appearance and identity} (opening keyframe and subject references), paired with timestamped action beats and explicit anti-drift constraints.
Coarse proxy synthesis reuses the same schema while locking the original action and stripping photorealistic shading.
Every synthesized candidate is audited side-by-side against its source recording by a VLM verifier across four axes---\emph{camera trajectory}, \emph{spatial layout}, \emph{subject motion}, and \emph{causal event timing}---retaining only consistent clips and retrying rejected renders.

\noindent\textbf{Chunk-level training prompt compilation (Fig.~\ref{fig:suppl-train-prompt}).}
For autoregressive DCL training, paired trajectories are sliced into temporally synchronized chunks and deterministically compiled into three-tier chunk prompts: a shared session style prefix, visibility-gated subject references admitted only when a subject appears in the current proxy, and chunk-local timestamped event descriptions.
Clip-wide global summaries are excluded from chunk prompts so no future events leak into earlier chunks.

\begin{table*}[t]
\centering
\scriptsize
\caption{\textbf{Blinded VLM (Gemini~3.8~Flash) Elo leaderboards on GWM-Bench Variation ($n=120$) and Solar-Bench Hard208 ($n=208$).} Axes assess perceptual clarity and coherence (\emph{Visual Quality}), proxy layout/trajectory alignment (\emph{Structure \& Motion}), first-frame style and subject-reference fidelity (\emph{Appearance}), chunk-synchronized event/subject control (\emph{Dynamic Inter.}, `---' on static-prompt Solar-Bench), and holistic preference (\emph{Overall}); mean Bradley--Terry Elo is $1000$~\cite{solarrenderer}. \textbf{Bold} marks the best score per step-regime group.}
\label{tab:suppl-elo}
\setlength{\tabcolsep}{4.5pt}
\begin{tabularx}{\textwidth}{@{}lXcccccc@{}}
\toprule
\textbf{Benchmark} & \textbf{Method} & \textbf{Steps} & \textbf{Visual Quality} & \textbf{Structure \& Motion} & \textbf{Appearance} & \textbf{Dynamic Inter.} & \textbf{Overall} \\
\midrule
\multirow{7}{*}{\shortstack[l]{\textbf{GWM-Bench}\\\textbf{Variation}\\($n=120$)}}
 & Streaming-CWM~\cite{liao2026streamingcwm} & 35 & 904 & 725 & 895 & 781 & 831 \\
 & SolarRenderer-AR~\cite{solarrenderer} & 30 & 992 & 1050 & 979 & 1016 & 998 \\
 & \textbf{ContextRenderer-33B (Ours)} & 30 & \textbf{1259} & \textbf{1156} & \textbf{1242} & \textbf{1193} & \textbf{1220} \\
\cmidrule(lr){2-8}
 & StreamRender-H3~\cite{streamrenderh3} & 2 & 470 & 777 & 602 & 783 & 687 \\
 & Streaming-CWM (4-step)~\cite{liao2026streamingcwm} & 4 & 927 & 857 & 916 & 851 & 892 \\
 & SolarRenderer-AR (4-step)~\cite{solarrenderer} & 4 & 1106 & 1185 & 1046 & 1075 & 1060 \\
 & \textbf{ContextRenderer-33B (Ours, 4-step)} & 4 & \textbf{1342} & \textbf{1250} & \textbf{1320} & \textbf{1302} & \textbf{1312} \\
\midrule
\multirow{7}{*}{\shortstack[l]{\textbf{Solar-Bench}\\\textbf{Hard208}\\($n=208$)}}
 & Streaming-CWM~\cite{liao2026streamingcwm} & 35 & 865 & 694 & 913 & --- & 782 \\
 & SolarRenderer-AR~\cite{solarrenderer} & 30 & 986 & \textbf{1078} & 1014 & --- & 1022 \\
 & \textbf{ContextRenderer-33B (Ours)} & 30 & \textbf{1103} & 1074 & \textbf{1127} & --- & \textbf{1127} \\
\cmidrule(lr){2-8}
 & StreamRender-H3~\cite{streamrenderh3} & 2 & 540 & 929 & 481 & --- & 603 \\
 & Streaming-CWM (4-step)~\cite{liao2026streamingcwm} & 4 & 1013 & 869 & 1032 & --- & 961 \\
 & SolarRenderer-AR (4-step)~\cite{solarrenderer} & 4 & 1219 & \textbf{1192} & 1183 & --- & 1232 \\
 & \textbf{ContextRenderer-33B (Ours, 4-step)} & 4 & \textbf{1274} & 1165 & \textbf{1250} & --- & \textbf{1273} \\
\bottomrule
\end{tabularx}
\end{table*}

\begin{table*}[t]
\centering
\scriptsize
\caption{\textbf{Extended automatic metrics on GWM-Bench Variation ($n=120$) and Solar-Bench Hard208 ($n=208$).} \emph{DOVER}~\cite{dover} and \emph{VideoScore2}~\cite{videoscore2} assess visual quality, \emph{Motion Err.}~\cite{cotracker3,solarrenderer} and \emph{Edge F1}~\cite{solarrenderer} measure CoTracker3 trajectory and proxy-contour alignment, and \emph{Adj.\ Sim.}~\cite{dinov2,solarrenderer} measures consecutive-frame DINOv2 similarity. \textbf{Bold} marks the best score per step-regime group; \texttt{N/A} marks Streaming-CWM freeze-frame collapse, and its proxy-motion neglect on Hard208 inflates \emph{Adj.\ Sim.}}
\label{tab:suppl-auto-var-hard}
\setlength{\tabcolsep}{4.5pt}
\begin{tabularx}{\textwidth}{@{}lXcccccc@{}}
\toprule
\textbf{Benchmark} & \textbf{Method} & \textbf{Steps} & \textbf{DOVER $\uparrow$} & \textbf{VideoScore2 $\uparrow$} & \textbf{Motion Err. $\downarrow$} & \textbf{Edge F1 $\uparrow$} & \textbf{Adj. Sim. $\uparrow$} \\
\midrule
\multirow{7}{*}{\shortstack[l]{\textbf{GWM-Bench}\\\textbf{Variation}\\($n=120$)}}
 & Streaming-CWM~\cite{liao2026streamingcwm} & 35 & \texttt{N/A} & \texttt{N/A} & 0.082 & 0.411 & 0.982 \\
 & SolarRenderer-AR~\cite{solarrenderer} & 30 & 0.477 & 3.133 & 0.055 & 0.417 & 0.975 \\
 & \textbf{ContextRenderer-33B (Ours)} & 30 & \textbf{0.527} & \textbf{3.318} & \textbf{0.040} & \textbf{0.469} & \textbf{0.983} \\
\cmidrule(lr){2-8}
 & StreamRender-H3~\cite{streamrenderh3} & 2 & 0.472 & 2.962 & 0.054 & 0.339 & 0.979 \\
 & Streaming-CWM (4-step)~\cite{liao2026streamingcwm} & 4 & \texttt{N/A} & \texttt{N/A} & 0.066 & 0.417 & 0.980 \\
 & SolarRenderer-AR (4-step)~\cite{solarrenderer} & 4 & 0.425 & 3.192 & 0.052 & 0.463 & 0.984 \\
 & \textbf{ContextRenderer-33B (Ours, 4-step)} & 4 & \textbf{0.519} & \textbf{3.284} & \textbf{0.029} & \textbf{0.504} & \textbf{0.985} \\
\midrule
\multirow{7}{*}{\shortstack[l]{\textbf{Solar-Bench}\\\textbf{Hard208}\\($n=208$)}}
 & Streaming-CWM~\cite{liao2026streamingcwm} & 35 & \texttt{N/A} & \texttt{N/A} & 0.092 & 0.187 & \textbf{0.980} \\
 & SolarRenderer-AR~\cite{solarrenderer} & 30 & 0.557 & 4.712 & \textbf{0.060} & \textbf{0.294} & 0.975 \\
 & \textbf{ContextRenderer-33B (Ours)} & 30 & \textbf{0.583} & \textbf{4.854} & 0.066 & 0.212 & \textbf{0.980} \\
\cmidrule(lr){2-8}
 & StreamRender-H3~\cite{streamrenderh3} & 2 & 0.520 & 4.319 & 0.064 & 0.213 & 0.971 \\
 & Streaming-CWM (4-step)~\cite{liao2026streamingcwm} & 4 & \texttt{N/A} & \texttt{N/A} & 0.087 & 0.183 & 0.977 \\
 & SolarRenderer-AR (4-step)~\cite{solarrenderer} & 4 & 0.565 & 4.775 & 0.068 & \textbf{0.335} & 0.977 \\
 & \textbf{ContextRenderer-33B (Ours, 4-step)} & 4 & \textbf{0.567} & \textbf{4.819} & \textbf{0.063} & 0.243 & \textbf{0.981} \\
\bottomrule
\end{tabularx}
\end{table*}

\subsection{GWM-Bench}
\label{sec:suppl-benchmarks}

GWM-Bench evaluates continuous interactive rendering across diverse proxy representations and changing world conditions on $320$ held-out clips ($124$ frames, $24$\,fps), strictly disjoint from WorldPrism training trajectories at the source-trajectory level. It is organized into two tasks:

\noindent\textbf{GWM-Bench Reconstruction ($200$ pairs).}
This split pairs real gameplay and open-world target recordings (covering AAA gameplay, open-world exploration, and Magpie~\cite{magpie} locomotion) with aligned coarse proxies spanning depth and semantic maps, low-poly and voxel renders, and untextured greybox geometry, testing faithful recovery of layout, motion, and appearance across proxy formats.

\noindent\textbf{GWM-Bench Variation ($120$ pairs).}
This split pairs multi-format proxies (depth and semantic maps, low-poly/voxel/capsule renders, and canonical RGB inputs) with stylized target references spanning dozens of distinct visual worlds (e.g., classical mythology, anime, ukiyo-e, ink wash, fantasy, and desert) across core gameplay interactions---entering and exiting vehicles, driving, combat attacks, and exploration. Every clip provides chunk-level dynamic prompts and visibility-gated subject references synchronized across autoregressive chunks.

\noindent\textbf{Dynamic-update hard subset ($n=63$).}
Used for component ablations (\mainref{main-paper Table~3}{Table~\ref{tab:ablation}}), this subset filters GWM-Bench Variation for sequences with newly introduced subjects or interaction-state updates, directly probing online condition following and cross-chunk continuity.

\section{Additional Experiments}
\label{sec:suppl-exp}

\begin{figure*}[t]
\centering
\includegraphics[width=\textwidth]{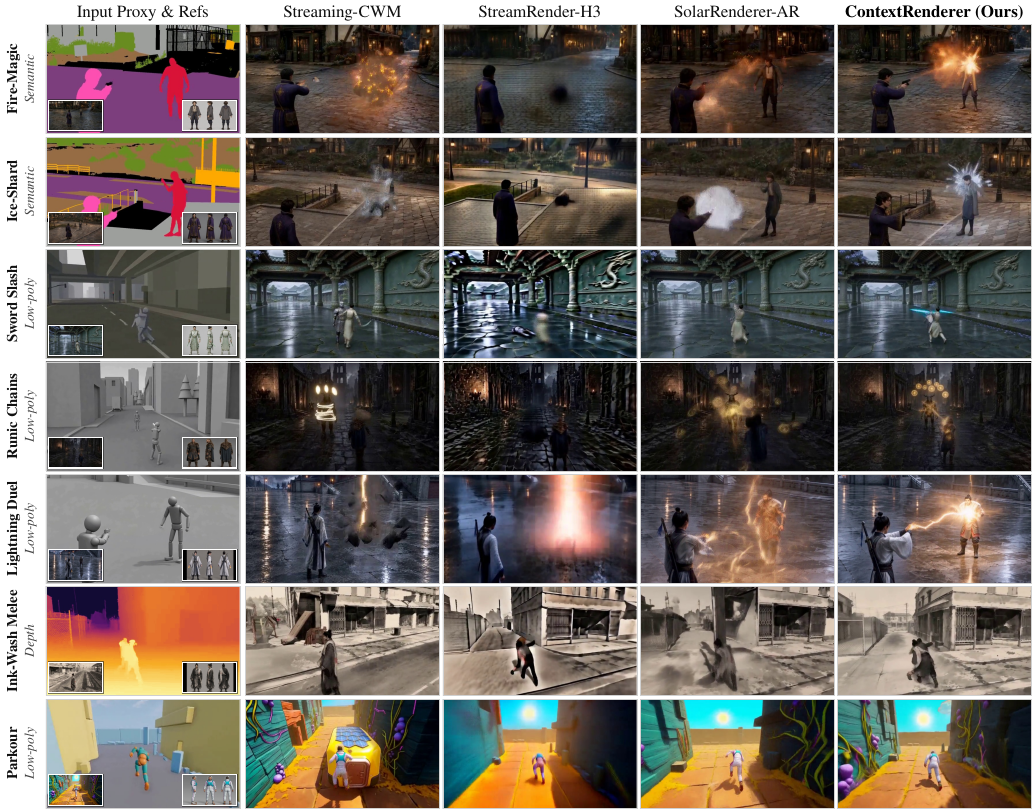}
\caption{\textbf{Multi-scene qualitative comparison on GWM-Bench Variation: magic VFX, combat, and parkour (33B autoregressive models).} Across seven scenarios spanning Semantic, Low-poly, and Depth proxies (target first frames in lower-left insets; subject references in lower-right insets), ContextRenderer renders crisp interaction VFX, wet-ground reflections, expressive ink-wash combat, and sharp high-motion traversal.}
\label{fig:suppl-vis-gwm}
\end{figure*}

\begin{figure*}[!t]
\centering
\includegraphics[width=\textwidth]{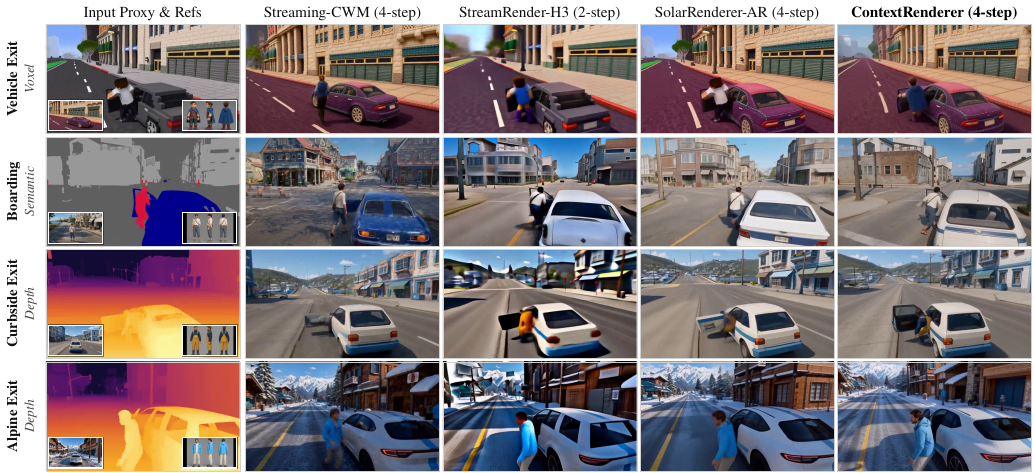}
\caption{\textbf{Qualitative comparison on GWM-Bench Variation: character--vehicle boarding and disembarking (few-step 33B models).} Across Voxel (\emph{Vehicle Exit}), Semantic (\emph{Boarding}), and Depth (\emph{Curbside Exit}, \emph{Alpine Exit}) proxies (first frames and subject references in insets), baselines suffer from mask leakage, disembarking blur, or vehicle warping, whereas ContextRenderer synchronizes door contact and binds emerging identities.}
\label{fig:suppl-vis-gwm-veh}
\vspace{\dblfloatsep}
\includegraphics[width=\textwidth]{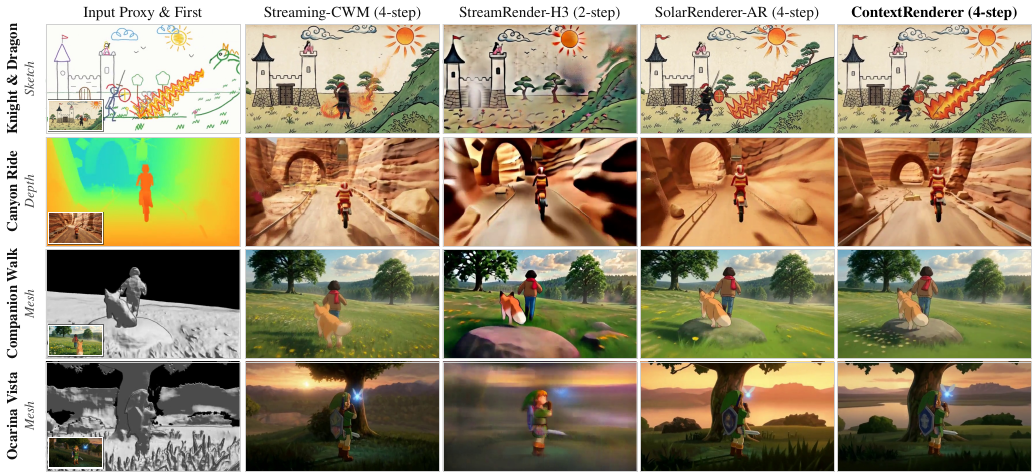}
\caption{\textbf{Cross-proxy qualitative comparison on Solar-Bench Hard208 (few-step 33B models).} Across unseen crayon sketches (\emph{Knight \& Dragon}), colorized depth maps (\emph{Canyon Ride}), and shaded meshes (\emph{Companion Walk}, \emph{Ocarina Vista}; first frames in insets), ContextRenderer preserves foreground subjects, scene geometry, and stylized lighting with higher visual clarity.}
\label{fig:suppl-vis-solar}
\end{figure*}

\begin{figure*}[!t]
\centering
\includegraphics[width=\textwidth]{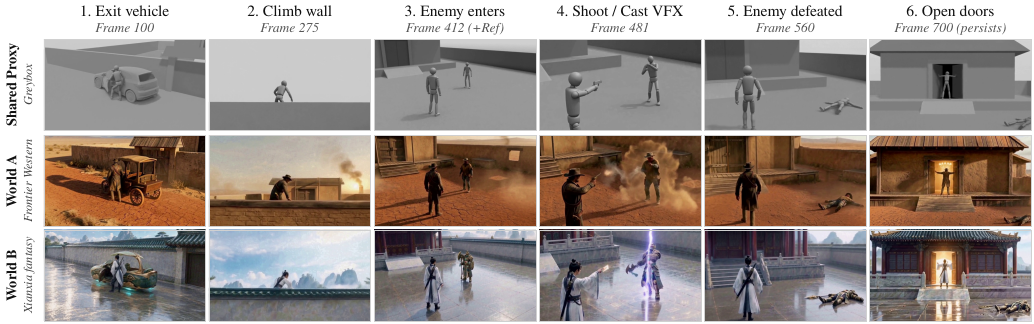}
\caption{\textbf{Long-horizon dual-world rendering from a shared greybox trajectory.} Conditioned on the same untextured greybox proxy (top), ContextRenderer renders a frontier Western courtyard (middle) and a wet-stone Xianxia courtyard (bottom) across vehicle exit, wall climbing, mid-clip enemy arrival (\texttt{+Ref}), world-specific VFX, and double-door opening while preserving the defeated opponent on the ground.}
\label{fig:suppl-long-dual}
\vspace{\dblfloatsep}
\includegraphics[width=\textwidth]{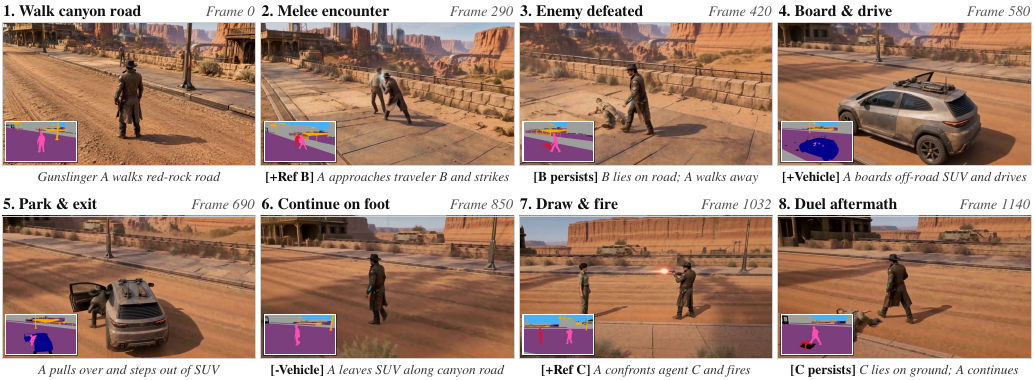}
\caption{\textbf{Extended 1000+-frame continuous interactive rendering session.} Complementing \mainref{main-paper Fig.~5}{Fig.~\ref{fig:qual-evolving}}, ContextRenderer sustains a multi-act Western rollout following protagonist A, driven by semantic proxies (lower-left insets) and chunk-level context updates across melee combat (\texttt{+Ref~B}), SUV boarding/driving/exiting (\texttt{+Vehicle}/\texttt{-Vehicle}), and a ranged showdown (\texttt{+Ref~C}) with persistent aftermath states.}
\label{fig:suppl-long-western}
\end{figure*}

\begin{figure*}[!t]
\centering
\includegraphics[width=\textwidth]{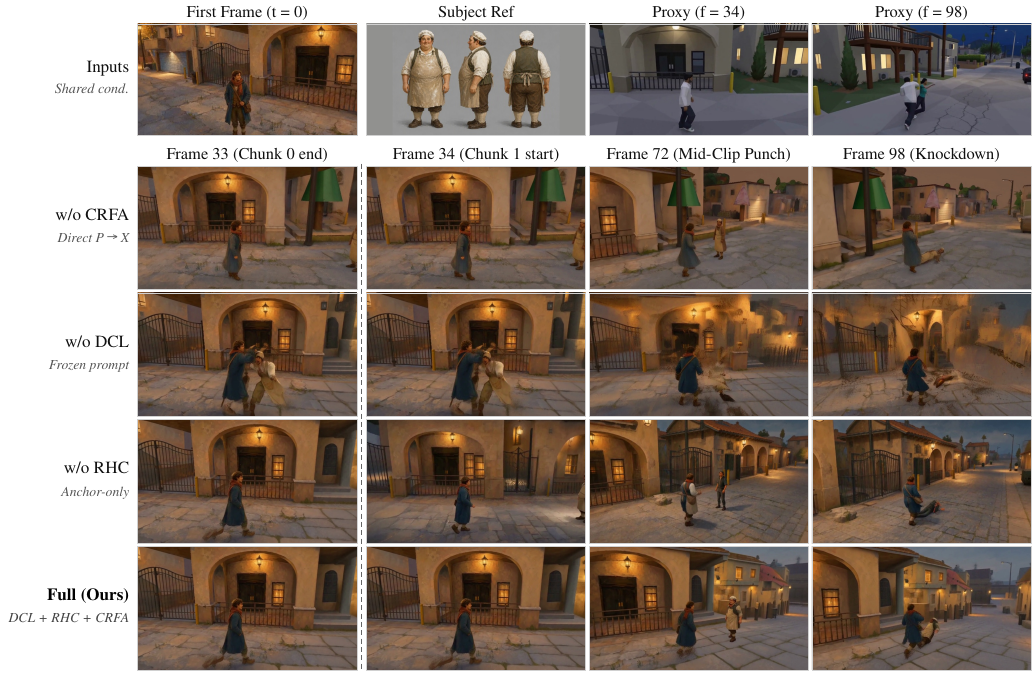}
\caption{\textbf{Qualitative component ablation corresponding to \mainref{main-paper Table~3}{Table~\ref{tab:ablation}}.} Top row: target first frame ($t=0$), newly introduced subject reference ($\mathcal{I}_k$), and Low-poly proxy keyframes; the vertical dashed line marks the chunk boundary (frames~33$\rightarrow$34). Omitting DCL, RHC, or CRFA respectively hallucinates premature combat and strike ghosting, introduces cross-chunk architectural jumps, or degrades subject structure and identity, whereas full ContextRenderer maintains seamless transitions, faithful identity binding, and exact timing.}
\label{fig:suppl-vis-ablation}
\end{figure*}

We expand on the evaluation in \mainref{Sec.~4 of the main paper}{Sec.~\ref{sec:experiments}} with blinded pairwise Elo ratings and few-step distillation analyses for the 33B models on visual variation, followed by cross-benchmark qualitative comparisons, extended long-horizon rollouts, and visual component ablations.

\subsection{Extended Quantitative Comparisons}
\label{sec:suppl-quant}

\noindent\textbf{Benchmark comparisons.}
Following~\cite{solarrenderer}, we use Gemini~3.8~Flash (\texttt{gemini-3.8-flash}, temperature~$0$) as the blinded VLM judge across all pairs of the seven multi-step (30--35 steps) and few-step (2--4 steps) models in both $(A,B)$ and $(B,A)$ orders, fitting a Bradley--Terry model per axis with mean Elo normalized to $1000$ (Table~\ref{tab:suppl-elo}), alongside the five automatic metrics from \mainref{main-paper Table~2}{Table~\ref{tab:main-variation}} (Table~\ref{tab:suppl-auto-var-hard}).
On GWM-Bench Variation, ContextRenderer ranks first on every VLM Elo axis and every automatic metric in both step regimes, outperforming the strongest baseline (SolarRenderer-AR) by $+222$ Overall Elo at 30 steps and $+252$ at 4 steps.
Its largest advantages appear in Visual Quality, Appearance, and Dynamic Interaction, reflecting faithful world-style rendering and chunk-synchronized subject/event control without sacrificing motion accuracy.
On the out-of-domain Solar-Bench Hard208, ContextRenderer generalizes zero-shot to unseen sketches, depth maps, and meshes, leading Overall Elo, perceptual quality (DOVER, VideoScore2), and temporal smoothness in both regimes while remaining competitive with SolarRenderer-AR on structure alignment.

\noindent\textbf{Few-step distillation.}
Across both benchmarks, 4-step decoupled DMD preserves perceptual fidelity on ContextRenderer (unlike SolarRenderer-AR's DOVER drop on GWM-Bench Variation) while matching or improving upon its 30-step teacher in Elo and proxy alignment~\cite{solarrenderer}.
By contrast, 2-step StreamRender-H3 suffers severe perceptual degradation, and Streaming-CWM exhibits frequent freeze-frame collapse or motion neglect (Table~\ref{tab:suppl-auto-var-hard}).

\subsection{Qualitative Comparisons Across Benchmarks}
\label{sec:suppl-qual}

\noindent\textbf{VFX, combat, and traversal.}
Fig.~\ref{fig:suppl-vis-gwm} compares multi-step renderings on GWM-Bench Variation across Semantic, Low-poly, and Depth proxies.
StreamRender-H3 suffers severe smearing and dark silhouettes across characters and spells (\emph{Fire-Magic}, \emph{Sword Slash}), while Streaming-CWM produces detached particle clouds, pose distortion, or proxy-tracking failure during fast motion (\emph{Parkour}).
SolarRenderer-AR maintains global scene structure but often renders flat or over-bloomed effects (e.g., flat overlays in \emph{Runic Chains} and blown-out highlights in \emph{Parkour}).
ContextRenderer generates physically grounded VFX that illuminate surrounding surfaces and wet-ground reflections (\emph{Lightning Duel}), alongside crisp ink-wash brushwork and sharp high-motion parkour.

\noindent\textbf{Character--vehicle interaction.}
Fig.~\ref{fig:suppl-vis-gwm-veh} compares the few-step models on vehicle boarding and disembarking across Voxel, Semantic, and Depth proxies, where emerging drivers are absent or occluded in the first frame.
StreamRender-H3 exhibits radial blur and melted vehicle geometry, while Streaming-CWM leaks blue segmentation masks onto the white hatchback (\emph{Boarding}) and blurs exiting drivers (\emph{Curbside Exit}, \emph{Alpine Exit}).
SolarRenderer-AR leaks voxel blocks onto the character (\emph{Vehicle Exit}) and blurs disembarking drivers into streaks (\emph{Curbside Exit}, \emph{Alpine Exit}), whereas ContextRenderer renders sharp, reference-consistent characters (e.g., the blue-caped figure in \emph{Vehicle Exit} and yellow overalls in \emph{Curbside Exit}) with crisp door-contact details.

\noindent\textbf{Out-of-domain generalization.}
On unseen Solar-Bench Hard208 structures---hand-drawn sketches, colorized depth maps, and shaded meshes (Fig.~\ref{fig:suppl-vis-solar})---StreamRender-H3 degrades into heavy motion blur and sky banding across all four scenes.
The 4-step baselines preserve overall scene geometry much better, yet Streaming-CWM exhibits flatter canyon lighting and obscured details (e.g., the dragon head lost in fire in \emph{Knight \& Dragon} and muted sunset tones in \emph{Ocarina Vista}), while SolarRenderer-AR shows softer rock shading and fine textures.
ContextRenderer delivers the sharpest stylized linework and dragon details in \emph{Knight \& Dragon}, high-contrast canyon lighting in \emph{Canyon Ride}, and crisp foliage and rim lighting in \emph{Companion Walk} and \emph{Ocarina Vista}.

\subsection{Extended Long-Horizon Interactive Rollouts}
\label{sec:suppl-long}

\noindent\textbf{Dual-world rendering.}
Fig.~\ref{fig:suppl-long-dual} demonstrates disentanglement between proxy motion/geometry and target world semantics over an unbroken long-horizon rollout.
Driven by the same untextured greybox sequence, ContextRenderer translates identical coarse geometry into either a dusty Western town or a misty Xianxia courtyard---adapting vehicles (wooden carriage vs.\ jade carriage), architecture, emerging enemy attire (\texttt{+Ref}), and action VFX (revolver smoke vs.\ purple lightning) while keeping the defeated opponent persistent on the ground through subsequent door-opening interactions.

\noindent\textbf{Multi-act rollout.}
Complementing \mainref{main-paper Fig.~5}{Fig.~\ref{fig:qual-evolving}}, Fig.~\ref{fig:suppl-long-western} extends continuous interactive rendering to 1000+ frames across open canyon terrain and sequential entity transitions.
Guided by semantic proxies and chunk-synchronized context updates, ContextRenderer maintains protagonist A's identity across 30+ autoregressive chunks while admitting and retiring auxiliary entities on schedule---binding traveler B for melee combat (\texttt{+Ref~B}), boarding and exiting an off-road SUV (\texttt{+Vehicle}/\texttt{-Vehicle}), and binding agent C for a ranged duel (\texttt{+Ref~C})---with persistent aftermath states and no cumulative color drift.

\subsection{Qualitative Component Ablation}
\label{sec:suppl-ablation}

Fig.~\ref{fig:suppl-vis-ablation} visualizes the failure modes behind \mainref{main-paper Table~3}{Table~\ref{tab:ablation}} across a chunk boundary (frames~33$\rightarrow$34) where a new subject enters with its reference image $\mathcal{I}_k$ prior to a melee knockdown.
Omitting CRFA weakens coarse-proxy grounding, degrading subject structure and failing to bind the arriving chef identity.
Freezing a clip-wide prompt (w/o DCL) leaks future combat into the solo-walking prefix---hallucinating a premature fight across frames~33--34---and collapses into ghosting during the actual strike.
Removing rolling history (w/o RHC) snaps architectural geometry and lighting across frames~33$\rightarrow$34 and loses opponent outfit consistency, whereas combining all three components yields seamless chunk transitions, faithful subject binding, and exact interaction timing.

\begin{figure*}[!t]
\centering
\fbox{\parbox{0.97\textwidth}{\footnotesize\ttfamily
\textbf{[Prompt A: Conditional Visual-Style Video Synthesis (High-Fidelity Variant)]}\\[2pt]
\textbf{subject\_definitions:} <Video 1> is the colorized semantic video (124 frames, 24 fps, 5.17s). It is the authority for camera trajectory, crop, subject slots, scale, and facing, without locking limb pose; never copy flat class colors or source weapons.\\
- static world: RGB(70,70,70) \quad - <Subject 1> (attacker): RGB(220,20,60) \quad - <Subject 2> (target): RGB(255,0,0) \quad - vehicle: RGB(0,0,142)\\
<Picture 1> is the high-fantasy opening keyframe (Frame 0): sunlit limestone city, emerald banners, golden afternoon light; <Subject 1> walks holding a two-handed greatsword low in both hands while <Subject 2> stands ahead. The hatchback becomes a carved wooden carriage.\\
<Picture 2> is the identity sheet for <Subject 1> only (same slot and heading; do not copy the studio backdrop or apply this identity to <Subject 2>).\\[2pt]
\textbf{summary:} Camera and spatial slots follow <Video 1>; environment and materials follow <Picture 1>. <Subject 1> approaches until 3.25s, executes a greatsword strike landing at 3.58s, and <Subject 2> falls at 3.75s.\\[2pt]
\textbf{retention\_analysis:} <Video 1>: partially\_preserved (camera, slots, and fall timing stay; source weapon is replaced). <Picture 1>: preserved (materials, lighting, and opening pose hold). <Subject 1/2>: partially\_preserved (world-native action and synchronized fall).\\[2pt]
\textbf{detailed\_description:}\\
- 0.00s--3.25s: <Subject 1> walks forward holding the solid greatsword low in both hands; <Subject 2> stands still.\\
- 3.25s--3.58s: <Subject 1> raises the greatsword overhead with both hands while feet remain grounded.\\
- At 3.58s: <Subject 1> chops downward onto <Subject 2>'s shoulder with contact impact (no glowing beam or laser).\\
- 3.75s--5.17s: <Subject 2> falls and stays down through the final frame; <Subject 1> holds the greatsword.\\
FAIL IF a beam connects the subjects; FAIL IF the scene remains the source game; FAIL IF impact occurs before 3.58s or <Subject 2> falls before 3.75s; FAIL IF <Subject 2> copies <Picture 2>.
}}
\vspace{3pt}

\fbox{\parbox{0.97\textwidth}{\footnotesize\ttfamily
\textbf{[Prompt B: Aligned Proxy Video Synthesis (Low-Poly Proxy Example)]}\\[2pt]
\textbf{subject\_definitions:} <Video 1> is the colorized semantic video (124 frames, 24 fps, 5.17s) and the sole authority for camera, crop, slots, scale, facing, and the original action. Never copy flat class colors.\\
<Picture 1> is the amateur low-poly opening of Frame 0: boxy faceless figures, flat vertex colors, coarse triangulated ground, faceted box buildings, no photorealistic textures. Preserve the original source action with no added weapons or magic.\\[2pt]
\textbf{detailed\_description:} Rebuild the entire shot in the low-poly proxy style matching <Picture 1>. Follow <Video 1>'s camera path and action timing exactly; <Subject 2> falls at 3.75s and stays down through 5.17s.\\
FAIL IF any sword, spell, or beam appears; FAIL IF characters revert to photorealistic shading; FAIL IF timing drifts from <Video 1>.
}}
\vspace{3pt}

\fbox{\parbox{0.97\textwidth}{\footnotesize\ttfamily
\textbf{[Prompt C: Automated VLM Quality-Control (QC) Verification Prompt]}\\[2pt]
Evaluate a conditional video synthesis pair. LEFT is the original source RGB video; RIGHT is the synthesized video (both 0.00s--5.17s; letterboxing is not a defect). The opening keyframe specifies appearance only; the identity sheet applies strictly to the designated subject.\\
The LEFT source video is the ground-truth authority for camera trajectory, spatial layout, subject motion, and causal event timing. An authorized stylized action may replace the source weapon/gesture in appearance, but MUST preserve its causal onset and impact timing.\\
Score four axes (Camera, Layout, Motion, Timing) with timestamped evidence and output \texttt{pass} or \texttt{reject} with failure tags.
}}
\caption{\textbf{Representative production prompts in the WorldPrism data engine.}
Top: structured synthesis prompt for a high-fidelity stylized realization.
Middle: aligned low-poly proxy synthesis prompt.
Bottom: side-by-side VLM verification prompt enforcing geometric and temporal fidelity.}
\label{fig:suppl-prompts}
\end{figure*}

\section{Prompt Templates}
\label{sec:suppl-prompt-templates}

Full production and training prompt strings referenced in Sec.~\ref{sec:suppl-prompt-qc} are shown in Figs.~\ref{fig:suppl-prompts} and~\ref{fig:suppl-train-prompt}.

\begin{figure*}[!t]
\centering
\fbox{\parbox{0.97\textwidth}{\small\ttfamily
\textbf{[Chunk-Level Training Prompt (MiniMax-H3, Continuation Chunk k=3)]}\\[3pt]
This is one shot. <Video 1> gives the proxy frames of the part being rendered now, and character sheets may be given for the subjects in it. When sources conflict: geometry, camera and motion follow <Video 1>; look follows <Picture 1>; subjects and events follow the chunk text.\\
\textbf{environment:} neoclassical 3D rendering with polished white marble, fluted stonework, and gilded bronze inlays; a wide paved stone concourse lined with classical colonnades, temple facades, and blue banners under a clear sky; bright direct sunlight casting sharp shadows across the stone ground.\\[3pt]
\textbf{subject\_definitions:}\\
<Picture 1> is the look reference of the whole shot: its materials, colours and lighting hold in every chunk.\\
<Video 1> is the proxy of the part being rendered now, seen from the same camera as the target video: it fixes the layout, where every subject is and how it moves, and the camera, frame by frame. Its rendering (materials, colours, lighting and level of detail) is not the target's.\\
<Picture 2> is the character sheet of <Subject 1>, the followed person, a man: a muscular young man with curly dark hair, wearing a bronze cuirass over an ivory tunic, a draped cream cloak, tall bronze greaves, and a golden laurel wreath; the sheet pose is an example.\\
<Picture 3> is the character sheet of <Subject 2>, the other person, a man: a slender young man with curly dark hair, wearing a sleeveless ivory chiton with a rope belt and bare feet; the sheet pose is an example.\\[3pt]
\textbf{summary:} This chunk continues the shot; <Picture 1> is not its current frame. <Subject 1>, the laurel-wreathed man in the centre, and <Subject 2>, the man in the ivory chiton, are visible in this chunk. [reference generation] The chunk follows the frame-by-frame structure and motion of <Video 1>. <Subject 1> lands a punch on <Subject 2> before coming to a halt; <Subject 2> collapses backward onto the stone ground.\\[3pt]
\textbf{retention\_analysis:}\\
<Picture 1>: partially\_preserved - materials, colours and lighting are kept in every chunk.\\
<Video 1>: attribute\_transfer - layout, subject positions and motion, and camera are followed; its rendering is not shown.\\
<Picture 2>: partially\_preserved - identity of <Subject 1> is kept; the sheet pose is an example.\\
<Picture 3>: partially\_preserved - identity of <Subject 2> is kept; the sheet pose is an example.\\[3pt]
\textbf{detailed\_description:} The target video is a neoclassical 3D rendering with polished white marble, fluted stonework, and gilded bronze inlays.\\
{}[Shot 1] <0.0 seconds> to <1.3 seconds>: <Subject 1>, the laurel-wreathed man, lands a punch on <Subject 2>, the man in the ivory chiton, before coming to a halt; <Subject 2> collapses backward onto the stone ground. At <0.5 seconds>, <Subject 1> stops moving. <Subject 2> lies on the ground.
}}
\caption{\textbf{Representative chunk-level training prompt compiled for MiniMax-H3.} A real continuation chunk ($k=3$) from WorldPrism combining the shared session prefix, visibility-gated subject references, and chunk-local proxy and event description fed to Qwen3-VL (empty audio placeholders omitted).}
\label{fig:suppl-train-prompt}
\end{figure*}

\end{document}